\documentclass[sn-apa]{sn-jnl}% APA Reference Style
\usepackage{CJKutf8}
\usepackage{graphicx}%
\usepackage{multirow}%
\usepackage{amsmath,amssymb,amsfonts}%
\usepackage{amsthm}%
\usepackage{mathrsfs}%
\usepackage[title]{appendix}%
\usepackage{xcolor}%
\usepackage{textcomp}%
\usepackage{manyfoot}%
\usepackage{booktabs}%
\usepackage{algorithm}%
\usepackage{algorithmicx}%
\usepackage{algpseudocode}%
\usepackage{listings}%
\usepackage{etoolbox}
\usepackage{multirow}
\usepackage{multicol}
\usepackage{setspace}
\usepackage{caption}
\usepackage{tikz}
\usepackage{amsmath}
\usepackage{array}
\usepackage{lineno}
\DeclareCaptionLabelSeparator*{spaced}{\\[2ex]}
\theoremstyle{thmstyleone}%
\usepackage{longtable}

\theoremstyle{thmstyletwo}%
\usepackage{booktabs}
\usepackage{appendix}

\theoremstyle{thmstylethree}%

\begin{document}
%\pagewiselinenumbers

\title{Fine-tuning ChatGPT for Automatic Scoring of Written Scientific Explanations in Chinese}


\author[1,4]{\fnm{Jie} \sur{Yang}}\email{yangjiebnuer@163.com}

\author[2,3]{\fnm{Ehsan} \sur{Latif}}\email{ehsan.latif@uga.edu}
\equalcont{These authors contributed equally to this work.}
\author[4]{\fnm{Yuze}
\sur{He}}\email{heyuze@mail.bnu.edu.cn}
\author*[2,3]{\fnm{Xiaoming} \sur{Zhai}}\email{xiaoming.zhai@uga.edu}
\equalcont{These authors contributed equally to this work.}

\affil[1]{\orgdiv{Faculty of Psychology}, \orgname{Beijing Normal University}, \orgaddress{ \city{Beijing}, \postcode{100875}, \state{Beijing}, \country{China}}}

\affil[2]{\orgdiv{AI4STEM Education Center}, \orgname{University of Georgia}, \orgaddress{ \city{Athens}, \postcode{30602}, \state{GA}, \country{USA}}}

\affil[3]{\orgdiv{Department of Mathematics, Science, and Social Studies Education}, \orgname{University of Georgia}, \orgaddress{ \city{Athens}, \postcode{30602}, \state{GA}, \country{USA}}}

\affil[4]{\orgdiv{Research institute of science education}, \orgname{Beijing Normal University}, \orgaddress{ \city{Beijing}, \postcode{100875}, \state{Beijing}, \country{China}}}


\abstract{
The development of explanations for scientific phenomena is crucial in science assessment. However, the scoring of students' written explanations is a challenging and resource-intensive process. Large language models (LLMs) have demonstrated the potential to address these challenges, particularly when the explanations are written in English, an alphabetic language. It remains unknown whether this approach can be applied to other logographic languages. This study thus explores the potential of fine-tuning ChatGPT, one advanced LLM, to automatically score scientific explanations written in Chinese. We collected and automatically scored student responses to seven scientific explanation tasks in Chinese, and examined the relationship between scoring accuracy and reasoning complexity with Kendall correlation. Finally, a qualitative analysis was conducted to explore how linguistic features influence scoring accuracy.  The results indicate that through domain-specific adaptation, the fine-tuned ChatGPT can accurately score students' written explanations in Chinese. However, scoring accuracy correlates with reasoning complexity, showing a negative correlation for lower-level responses and a positive one for higher-level responses. The model tends to overrate complex reasoning for low-level responses with  complex sentence structures and underrate high-level responses, using generalizing, summarizing, or simple causal reasoning. These opposing correlations are linked to different linguistic features. The comprehensiveness of student responses is often in tension with the simplicity and clarity of language structure in terms of scoring accuracy. For lower-level responses, simplicity and clarity are prioritized, leading to more accurate scores for simpler and shorter responses. For higher-level responses, comprehensiveness is prioritized, resulting in more accurate scores for long and information-rich responses.
These findings demonstrate the effectiveness of LLMs in automatic scoring within a Chinese context and highlight the importance of considering linguistic features and reasoning complexity in developing and fine-tuning automatic scoring models for educational assessments.
}




\maketitle

\section{Introduction}\label{background; research gap; research foucse}

Constructing scientific explanations for real-world phenomena is fundamental in scientific inquiry and crucial for educational practices. This practice goes beyond descriptions of natural patterns and requires students to explain how or why a phenomenon occurs by seeking evidence and using scientific knowledge \citep{mcneill2008inquiry,driver2000establishing}. Such activities reflect students' scientific literacy and deepen their understanding of science \citep{Yao2018explanation}. Consequently, national and regional science curriculum standards from countries such as the United States, Canada, Australia, and China emphasize “constructing scientific explanations” as a key scientific practice for students. Educational research and international large-scale assessments, such as PISA and NAEP, also employ scientific explanation as a primary means to evaluate students’ science literacy\citep{OECD2019,zhai2023large}. These tasks typically present students with a natural phenomenon or experimental results, requiring them to write coherent explanations.  While these tasks offer a relatively comprehensive and deep assessment of students' abilities, scoring written explanations presents significant challenges. The writing's complex structure and diverse connotations demand considerable time and effort to score. Additionally, the intricate scoring criteria and the high costs associated with them can impede their practical application in everyday teaching.
          
With the development of artificial intelligence (AI), automatic scoring and developing learning environments for the automated analysis of learning have emerged as one of the most dynamic and cutting-edge directions in science education\citep{kubsch2022toward,mislevy2020automated,zhai2024ai}. This technology alleviates the considerable time and effort teachers traditionally dedicate to scoring, enabling the provision of immediate feedback, which facilitates precise instructional decision-making and personalized learning experiences\citep{lamb2021computational, riordan2020empirical}. This integration of technology represents a transformative shift in education, highlighting the extensive potential of automated scoring systems.  Recently AI and machine learning have been employed to score scientific explanations,  yielding varying accuracy\citep{zhai2021meta}. Notably, ChatGPT has demonstrated significant potential in this area\citep{latif2024fine}, underscoring the growing impact of AI on educational practices, particularly in the objective and efficient assessment of complex student responses.

However, existing studies mainly focus on English texts \citep{ariely2023machine}, leaving a significant gap in our understanding of other languages, such as Chinese. As a logographic language, Chinese is marked by distinct linguistic characteristics and cultural nuances that differ from alphabetic languages. For example, Chinese speakers often employ analytical thinking, and thus reasoning in Chinese written communication can be implicit, and with frequent use of idioms\citep{wang2013differences, Wang2021automated}. Evidence indicates that the character and word features were crucial for LLM to score students' constructed responses\citep{gombert2023coding}. Additionally, linguistic diversity could lead students to construct and articulate their scientific reasoning in ways that diverge significantly from those of English-speaking students\citep{Williamsf2010Chinese}. studies have shown that the  reasoning ability of LLMs varies by tasks\citep{dhar2023we, John2024GPTreasoning}, and there is a notable decline in model performance as the complexity and diversity of the tasks increase \citep{haudek2023examining}. ChatGPT, for example, faces constraints in solving reasoning problems, particularly those involving multi-step reasoning\citep{zhang2023assessing}.  It is thus valuable to learn how the challenges that ChatGPT and its variants may encounter in comprehending students’ written explanations with complex reasoning, a common requirement in scientific explanations.

To fill the existing gaps, we collected 7626 scientific explanations written in Chinese by middle school students across seven science tasks. We then used the responses to fine-tune ChatGPT and examined how the scoring accuracy vary by the degree of reasoning complexity and students' performance level. This study answers the following research questions:
\begin{enumerate}
    \item How accurate is Fine-tuned ChatGPT in scoring written scientific explanations in Chinese?
    \item To what degree is reasoning complexity in student writing associated with the scoring accuracy of Fine-tuned ChatGPT?
    \item How do linguistic features of students' written responses account for the scoring performance of Fine-tuned ChatGPT?
\end{enumerate}

\section{Automatic Scoring of Written Scientific Explanations}

Scientific explanation involves the construction and communication of valid claims based on evidence and reasoning, typically requiring multiple reasoning steps. For example, explaining the causes and effects of climate change using scientific evidence and models or predicting the outcome of a chemical reaction based on the conservation of mass and the periodic table, demands complex reasoning \citep{krell2022scientific}. Recognized as a crucial skill in science education, scientific explanation supports cognitive, affective, and social aspects of science learning, and helps develop students' conceptual understanding \citep{de2019constructing}, inquiry skills, and other scientific literacy as they examine data, connect ideas, justify their claims build  \citep{mcneill2008inquiry, reiser2012engaging}. Furthermore, engaging in scientific explanations immerses students in authentic scientific practices, such as generating hypotheses, designing experiments, analyzing data, and drawing conclusions \citep{Vosniadou2019understanding}. These practices foster curiosity, wonder, and critical thinking \citep{Bjerknes2024curiosity}. Given its significance, many countries have integrated scientific explanation as a critical curriculum goal within their science standards, such as the Next Generation Science Standards (NGSS) in the United States, the National Curriculum: Science in England, %\textcolor{blue}{
the Australian Curriculum: Science
%}
as well as China's science curriculum standards all emphasize the importance of students being able to explain natural phenomena. 

To facilitate scientific explanation practices in the classroom, researchers have leveraged computer technology to automatically score students' responses. This technology can offer timely and personalized feedback and reduce teachers' workload. Common techniques for automatic scoring include machine learning, natural language processing, or rule-based logic, which are applied to analyze and score written scientific explanations. For example, latent semantic analysis has been used to evaluate undergraduate majors' and non-majors' written explanations of evolutionary change with high accuracy \citep{Ha2011computerized, Nehm2012Transforming, Moharreri2014EvoGrader}. Similarly, \cite{Liqutomated} employed a combination of rule-based methods to extract and score components of scientific explanation from student responses in an online science inquiry system, demonstrating a strong correlation between automated and human scores across various tasks and questions. However, individual models may have limitations in accuracy, robustness, and generalization. To address these issues, some researchers apply ensemble methods to combine the strengths of different models and reduce the errors or biases of individual models in automatic scoring. For example, \cite{ormerod2023automated} introduced a short answer scoring engine composed of an ensemble of deep neural networks and a latent semantic analysis-based model to score student responses in a national assessment program. Their finding demonstrated that the engine achieved above-human-level accuracy and could also provide high-level explanations for the scores.

Recently, Large Language Models (LLMs), such as BERT, GPT, and XLNet have revolutionized natural language processing, and are regarded as a significant advancement in automatic scoring. LLMs are neural network architectures capable of learning from large amounts of text data and generating fluent and coherent texts. These models can also capture the semantic and syntactic features of natural language, including word choice, grammar, style, and coherence to automatically score students’ responses. Studies have shown that LLMs outperform other automatic scoring methods, such as semantic distance scoring \citep{beaty2021automating}, and more effectively capture the nuances and diversity of natural language, resulting in a closer correlation with human judgments of response quality. These models can be fine-tuned on specific domains or tasks to achieve state-of-the-art results in various natural language processing applications, including automatic scoring of written scientific explanations. For example, \cite{kumar2021automated} used both deep neural networks and state-of-the-art NLP tools to predict finer-grained rubric scores for student essays and identify key features of an effective rubric scoring model. They also demonstrated how LLMs can provide interpretable and transparent feedback for the scores.

One of the most advanced LLMs is ChatGPT, developed by OpenAI. It is designed to simulate conversations with human users and perform various tasks based on natural language inputs. ChatGPT can also be used for automatic scoring of written scientific explanations, as it can understand and generate natural language responses based on the CER framework or other criteria \citep{zhai2023chatgpt}. For example, \cite{latif2024fine} fine-tuned ChatGPT on six assessment tasks using a diverse dataset of middle and high school student responses along with expert scoring. They compared the performance of fine-tuned ChatGPT with that of the fine-tuned state-of-the-art language model, BERT, developed by Google. The results demonstrated that fine-tuned ChatGPT achieved significantly higher scoring accuracy than BERT across all the tasks and also can provide explanations for the scores. ChatGPT and other LLMs hold great potential for the automatic scoring of written scientific explanations, offering accurate, reliable, and interpretable scores and feedback for student responses \cite{lee2023applying}. However, several challenges and limitations need to be addressed. These include ethical concerns associated with the use of these models, the quality and diversity of the data used for fine-tuning, and the generalizability and validity of the scores across different domains and contexts. More empirical research is needed to focus on these aspects and explore the best practices and standards for using LLMs for the automatic scoring of written scientific explanations.

\begin{CJK*}{UTF8}{gbsn}

\section{Students' Writing of Science in Chinese}

The research has found that writing scientific explanations in both Chinese and English requires complex mental processes expressed with words and sentences, but these processes differ greatly \citep{liao2012effectivenesss}, reflecting the languish features of Chinese and English. Chinese is a widely used language worldwide, but it exhibits significant differences from English in scientific writing.  Chinese writing tends to be more circular, visual, and synthetic, whereas English writing tends to adopt linear, rational, and analytical style\citep{chen2021influence,sun2017cultural}. For instance, in describing locations, Chinese narratives typically start with the largest geographical entity and proceed to smaller ones, while English is the opposite\citep{wang2013differences}. Another difference is that English relies more on sentence structure, whereas Chinese writing conveys meanings more through word selection. English sentences typically feature an explicit subject-predicate structure, making word recognition easier, while Chinese sentences do not always adhere to this pattern clearly, making their sentence structures less explicit \citep{yang2018subword}. So, implicit expression is considered a more advanced way of expression in Chinese writing,  while explicit expression is preferred in English writing. 

These linguistic distinctions were also reflected in the vocabulary and thinking patterns. Chinese people prefer comprehensive thinking, using words that reflect holistic concepts, in contrast to the more segmented, logical vocabulary prevalent in English writing. For example, the Chinese character 手(hand) is a holistic concept, fingers (手指), wrists (手腕), and fists (拳头) are all parts of the hand. This holistic perspective is evident in phrases like``\textit{手}"上戴的表 (watch on his \textit{``wrist}") and \textit{``手"}上戴的戒指 (ring on his \textit{``finger"})\citep{wang2013differences}. Moreover, the Chinese written language, influenced by both classical and vernacular forms, features a wealth of unique synonyms. For instance, 眼 and 目 both refer to the organ of sight, but they convey slightly different meanings and are used differently. Additionally, some synonyms in Chinese may not have direct equivalents in other languages. For example, the Chinese word 笑 encompasses different manners of laughing, but in English, there are many different words to express all kinds of ways to laugh, such as giggle (咯咯出声笑)， sneer(冷笑)， smile(微笑)， chortle(哈哈大笑),  snigger(暗笑), all of these words are translated into Chinese should use the form like this: ``manner +笑"\citep{chen2021influence}. 

Thus, the automated scoring of written responses in Chinese could be more difficult than in English, especially for complex reasoning responses, due to the richness in vocabulary and sentence patterns in Chinese\citep{li2019word}.

\section{Methods}
\subsection{Participants}
Participants were recruited through a stratified sampling method from schools in Beijing, ensuring a diverse representation of different school types, including municipal-level demonstration schools, district-level demonstration schools, and regular schools. School administrators communicated the study details to students and their guardians, ensuring that participants were explicitly informed about the research and voluntarily agreed to participate. This study selected 1,593 students in grades 8-12 as participants, stratified by school level, including 872 high school students, accounting for 54.7\%, and 721 middle school students, accounting for 45.3\%. Of the total population, 830 are boys, accounting for 52.1\%, and 763 are girls, accounting for 47.9\%.

\subsection{Instruments}\label{overview intro of tasks, }
We selected seven scientific explanation items from our established assessment bank to construct a scientific explanation ability test instrument. The selected items demonstrate high reliability and validity, and they have been extensively employed in prior studies, such as \cite{Yao2018explanation} and \cite{zhai2022assessing}. Each item in the test was presented as open-ended questions, beginning with a detailed description of a problem scenario and phenomena to provide students with the necessary context for their explanations. Students were asked to explain observed phenomena or predict phenomena based on the given information. Students need to organize their language to construct explanations to answer the questions.  Table \ref{Instruments} provides an overview of the instrument used. The instrument includes a mix of scenarios: four out of 7 items are led by daily life scenarios, while the remaining three involve scientific lab contexts. The seven items include both direct explanation and prediction plus explanation: four items explicitly present phenomena that students are required to explain, while the other three require students to predict and explain potential phenomena that could arise from the described scenarios.
\begin{table}[htbp]
\centering
\caption{\textit{Overview of the Instruments}}
\label{Instruments}
\begin{tabular}{p{0.04\linewidth} p{0.1\linewidth} p{0.1\linewidth} p{0.65\linewidth}} 
\hline 
\textbf{Item} & \textbf{Item Scenario} & \textbf{Item Task} & \textbf{Description} \\ 
\hline  
Item 1 & daily life scenarios & explain phenomena & Ice can only cool water, not cause it to boil. However, under special conditions, ice can indeed make water boil. To demonstrate this, fill a flask halfway with water and heat it over a flame until the water boils. Once boiling, remove the flask from the heat and seal the mouth of the flask with a stopper, causing the boiling to stop. If you then invert the flask and place some crushed ice on the bottom, you will immediately see the water start to boil again. Explain the reason for this phenomenon. \\ 
Item 2 & lab contexts & predict phenomena & Imagine you are in deep space with a large rock at rest relative to you. You gently push it. Please describe in detail the motion of both you and the rock during and after the push, and explain the reason behind this motion. \\ 
Item 3 & lab contexts & predict phenomena & A bar magnet with its N pole facing downwards and its S pole facing upwards is freely falling from a nearby height along the axis of a coil positioned. As the magnet approaches the coil, what effect will the coil have on the magnet? Explain the phenomena from the perspective of energy conservation. \\ 
Item 4 & daily life scenarios & explain phenomena & When using a pump to inflate a bicycle tire, the pump becomes warm. Explain the primary reason for this. \\ 
Item 5 & daily life scenarios & predict phenomena & Predict the eventual fate of colorful hydrogen balloons released into the sky during festivals, and explain. \\ 
Item 6 & lab contexts & explain phenomena & Some people say, "If we could lower the temperature of all the seawater on Earth by 0.1°C, the released internal energy could effectively alleviate the energy crisis." Is this proposal feasible? Please explain the reason. \\ 
Item 7 & daily life scenarios & explain phenomena & When a drop of ink is dropped into water, the ink rapidly disperses throughout the water. The rate of dispersion is greater in hot water than in cold water. Please explain this phenomenon. \\ 
\hline 
\end{tabular}
\end{table}

\vspace{0.5cm}

Here, Item 3 ``\textit{Magnet fall task}" and Item 7 ``\textit{Ink diffusion task}" are used as examples for specific explanation (see Figure 1). Item 3 is set in a scientific lab context, ``\textit{a bar magnet with its N pole facing downwards and its S pole facing upwards is freely falling from a nearby height along the axis of a coil positioned.}" The scenario outlines two crucial pieces of information needed for explanation: (1) the magnet has a certain speed; (2) the magnet and the coil form a system. Then the question follows, ``\textit{As the magnet approaches the coil, what effect will the coil have on the magnet? Explain the phenomena from the perspective of energy conservation.}" This asks students to use the information implied in the context, coupled with the law of energy conservation to reason and make predictions about the phenomenon. However, Item 7 is set in a daily life scenario, ``\textit{When a drop of ink is dropped into water, the ink rapidly disperses throughout the water. The rate of dispersion is greater in hot water than in cold water. }" The scenario outlines one crucial information needed for explanation: The temperature of hot water is higher than that of cold water. Then the requirement follows, ``\textit{Please explain this phenomenon.}" This requires students to use the information implied in the situation coupled with the molecular diffusion theories to reason and explain the phenomenon.

All the items are  developed based on the Chinese curriculum standards, with students expected to meet the corresponding learning expectations. Items 1, 4, 6, and 7 align with middle school standards. Items 2, 3, and 5 are aligned with high school standards. The consistency of the items has been verified in previous studies, such as \cite{Yao2018explanation}. To ensure that students had the opportunity to learn the relevant content before taking the assessment, middle school students completed items aligned with their grade level, and high school students were tested on all seven items. Assessments were performed as paper-and-pencil tests in classroom settings. Students completed tasks related to scientific explanations under supervised conditions to ensure consistency across the different schools. Notably, the testing time was not restricted, and students could completely engage with the test items without time pressure. 

 \begin{figure}
    \centering
    \includegraphics[width=1\linewidth]{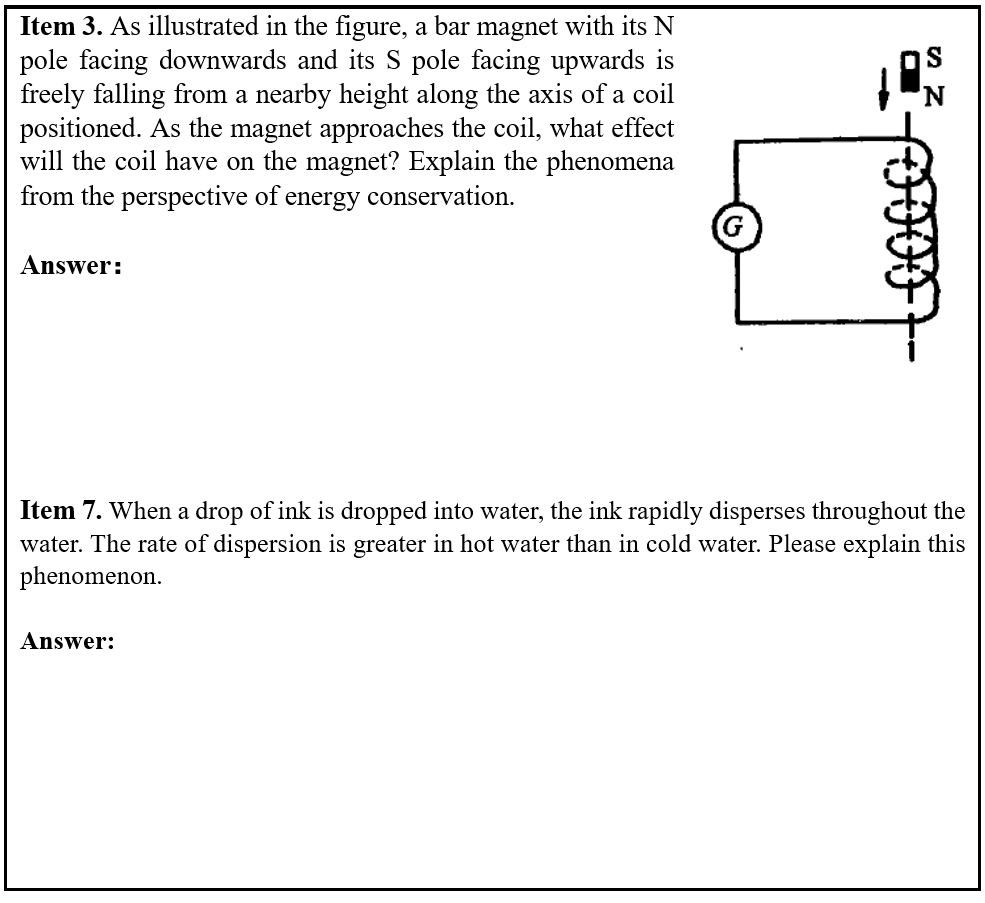}
    \caption{Exemplar scientific explanation items}
    \label{fig:enter-label}
\end{figure}

\subsection{Human Scoring}
The first stage is preprocessing the collected student responses. In this study, students completed their answers on paper, so we first eliminated unclear student responses, and unrecognizable fonts or blank questionnaires. To facilitate further analysis, we transcribed the remaining student responses from Chinese handwritten texts into print text. The final number of samples retained and transcribed for each item is shown in Table~\ref{tab:datasets}.

\begin{table}[htbp]
\centering
\caption{\textit{Overview of the Datasets Used in the Study}}
\label{tab:datasets}
\begin{tabular}{llccc}
\hline
\textbf{Datasets} & \textbf{Category} & \textbf{No. of Labels} & \textbf{Training Samples} & \textbf{Testing Samples} \\ \hline
Item 1 & Multi-Label/Multi-Class & 4 & 964 & 242 \\ 
Item 2 & Multi-Label/Multi-Class & 5 & 647 & 162 \\ 
Item 3 & Multi-Label/Multi-Class & 5 & 668 & 167 \\ 
Item 4 & Multi-Label/Multi-Class & 4 & 1011 & 253 \\ 
Item 5 & Multi-Label/Multi-Class & 5 & 630 & 158 \\ 
Item 6 & Multi-Label/Multi-Class & 4 & 1317 & 330 \\ 
Item 7 & Multi-Label/Multi-Class & 4 & 861 & 216 \\ \hline
\end{tabular}
\vspace{0.5cm}
\end{table}

\vspace{0.5cm}

The second stage employed human raters to score the transcribed text according to the scoring criteria(Table \ref{Rubrics}). The scoring criteria include two parts: Holistic scoring rubrics and Analytic scoring rubrics . The Holistic scoring rubrics are consistent with the evaluation methods used in daily teaching, that is, to evaluate the overall performance of students in answering each question. The overall scoring criteria are divided into two categories: correct explanation, and incorrect explanation. Students’ answers are scored as 1 and 0, respectively. 
The analytic scoring rubrics score student’s explanations corresponding to the four elements of data, theory, reasoning, and phenomena separately, according to the ``Phenomenon-Theory-Data-Reasoning (PTDR)” framework \citep{Yao2018explanation}. When the explanation constructed by students correctly displays a certain explanation element, that element is scored as 1, otherwise it is scored as 0. Analytical rubrics allow for more precise identification of specific elements in student responses, reducing potential biases and increasing scoring consistency compared to multi-level holistic scoring\citep{Wang2021automated}. It should be noted that 3 out of 7 items provide phenomena. For these items, students were not asked to identify the phenomena, and the Phenomenon element was not scored.

\begin{table}[h]
\centering
\caption{\textit{Rubrics of Magnet Fall Task}}
\label{Rubrics}
\begin{tabular}{>{\centering\arraybackslash}p{2.5cm} >{\centering\arraybackslash}p{1cm} p{8cm}} 
\hline  
\textbf{Explanation Element} & \textbf{Point} & \textbf{Description} \\ 
\hline  
Holistic scoring & 1 & Be able to view the magnet and coil as a system, pointing out that the kinetic energy of the magnet is converted into electromagnetic energy, and determine that the coil has a repulsive force on the falling of the magnet. \\ 
 & 0 & An incorrect claim about phenomena or a correct claim with a wrong explanation/extremely incomplete explanation.\\ 
\hline  
Data element of analytic scoring & 1 & Data include: the magnet has speed, and constitutes a system with the coil. \\ 
 & 0 & Other data or no data mentioned. \\ 
\hline  
Theory element of analytic scoring & 1 & Use the law of conservation of energy to explain the phenomenon. \\ 
 & 0 & Incorrect theory or no theory invoked. \\ 
\hline  
Reasoning element of analytic scoring & 1 & In the reasoning process the data, theory, and phenomenon are consistent. \\ 
 & 0 & Incorrect reasoning or no reasoning. \\ 
\hline  
Phenomenon element of analytic scoring & 1 & As the magnet approaches the coil, the coil exerts a repulsive force on it. \\ 
 & 0 & Incorrect claim about phenomena or no claim. \\ 
\hline
\end{tabular}
\end{table}

The human scoring results form the foundational data for developing and validating the algorithmic models. To obtain a better performance model, high inter-rater consistency (human-human consistency) is necessary. Four graduate students majoring in science education completed the scoring process. All of them possess content expertise relevant to the assessment items and were well-equipped to accurately interpret students' responses in alignment with the intended content goals of the assessment. We provided the raters with professional learning on how to score students’ responses with rubrics. They were then paired to score a sample of 50 randomly selected responses. If the human-human consistency, measured by Cohen's Kappa, was below 0.8, the raters discussed the discrepancies and re-scored another sample of 50 responses until a Cohen Kappa over 0.8 was achieved. Following this,  the raters scored the remaining student responses to the item independently. 

\subsection{Data Analysis}
\subsubsection{Fine-tuning ChatGPT}\label{Ehsan}
LLMs like ChatGPT have shown great potential because they can handle many different uses in general. However, it is important to fine-tune these models using specialized multi-lingual (e.g., Chinese) datasets for specific areas like education. This is especially true when computers automatically check how well students answer questions. In this study, we explain how to finetune ChatGPT to score complex written responses of students. Our dataset (see Table~\ref{tab:datasets}) consists of student-written responses and human-expert grading for holistic and analytical classes with 5 scoring perspectives. Before fine-tuning the ChatGPT using OpenAI service, it is crucial to process the data based on the OpenAI guidelines\footnote{\url{https://platform.openai.com/docs/guides/fine-tuning/preparing-your-dataset}} as follows:

\textbf{Data Cleaning and Privacy Assurance:} The first stage was going over the dataset in great detail to find and eliminate any information that wasn't directly related to the study's goals. More crucially, this step was essential to protecting the pupils' privacy and confidentiality. All personally identifying information, including names and unique identifiers, was painstakingly deleted. This action was taken to preserve ethical norms, protect the privacy of each individual whose data was used in this study, and ensure that the data was clean.

\textbf{Tokenization and Language Adaptation:} The students' textual responses had to be carefully converted into a format that the model could understand, as the language structures of Chinese are somewhat complicated. This was a complex process of translating the rich subtleties of the Chinese language into a digital format that preserves the breadth and context of the student's remarks rather than just a technical tokenization approach. Special care was taken to ensure that the intricacies inherent in the Chinese language, such as idioms and implicit reasoning, were not overlooked during the tokenization process.

\textbf{Data Standardization and Uploading:} The dataset was subjected to a standardization process following tokenization and cleaning. This puts the data in an OpenAI platform-compatible structured format, usually JSON. Then, using the file upload API\footnote{\url{https://platform.openai.com/docs/api-reference/files}}, the standardized files were safely uploaded to the OpenAI server. This was an important step since it took our carefully prepared dataset and put it into the AI processing domain, which started the process of turning raw data into meaningful conclusions.

\textbf{Anonymization for Ethical Compliance:} The data was further anonymized after identifying information was eliminated. Random codes were used in place of any possible identifiers during this procedure. This step maintained a high quality of ethical research practice by ensuring that even if someone were to acquire the data, it would be impossible to relate it back to specific students.

\textbf{Quality Checks and Balancing:} To ensure the dataset was representative and intact, it went through several quality checks. This involved distributing the dataset among various student answer durations and reasoning complexity levels. The goal was to produce a clean, well-structured dataset that accurately represented various student skills and cognitive processes.

Given the unique challenges of scoring scientific explanations in Chinese, fine-tuning ChatGPT was conducted with specific considerations:

\textbf{Loss Function:} A regression-based loss function was used for continuous scores and categorical scores, and a classification loss was used. The intricate organization of scientific explanations, where scores frequently fall into complex categories or necessitate a scale that indicates differing degrees of accuracy and completeness, impacted this decision.

\textbf{Learning Rate:} A cautiously low learning rate was our starting point. Because Chinese has distinct syntactic and semantic features that set it apart from the primarily English corpus, the model was initially trained on, this was important. Periodic assessments implemented gradual modifications in the learning rate to provide the best possible adaptation while retaining the model's original strengths.

\textbf{Epochs:} It was vital to use multiple epochs, particularly given the complexity of scientific reasoning in the student explanations. Because reasoning in Chinese texts is implicit and context-dependent, it was important to control validation loss to avoid overfitting carefully and instead learn sophisticated reasoning patterns.

\textbf{Batch Size:} To meet the computational requirements of processing Chinese characters and the intricate structures of scientific explanations, the batch size was optimized. The selected size allowed for effective processing without sacrificing the caliber of the training.

\textbf{Data Augmentation:} More data augmentation methods were applied to address the problem of multi-step reasoning in Chinese texts. To improve the model's comprehension and scoring of such responses, artificially generated samples that imitate complicated reasoning processes were incorporated.

Post-fine-tuning, the model was evaluated and validated using a set distinct from the training data. The evaluation involved:

\textbf{Metrics:} We used metrics such as Mean Absolute Error (MAE) or classification accuracy to gauge the model’s scoring precision against human raters. This step is essential to assess the effectiveness of the fine-tuning process.

\textbf{Comparison with Baseline:} The performance of the fine-tuned model was compared with the original GPT-3.5-turbo model. This comparison helps quantify the benefits of domain-specific fine-tuning and understanding the improvements in the model's ability to score student responses.

\subsubsection{Alignment with Research Objectives:}
The fine-tuning process was meticulously aligned with the research objectives and the questions raised in the introduction:

\textbf{ Scoring Accuracy:} The fine-tuning procedure was designed to consider the subtleties of scientific language and reasoning in student writings to answer the first research question about the model's accuracy in evaluating written scientific explanations in Chinese. The model's performance was continuously assessed against various explanations to guarantee high accuracy. Scoring accuracy refers to the agreement between machine scoring and human scoring\citep{zhai2021meta}. For example, an accuracy of 80\% indicates that the machine scores align with human scores for 80\% of the student samples. 

\textbf{Reasoning Complexity:} Special focus was paid to fine-tuning the model to distinguish different levels of reasoning difficulty in response to the second and third questions regarding the effect of reasoning complexity on scoring accuracy. As part of this, the model was trained to recognize implicit reasoning processes and context-dependent explanations prevalent in Chinese texts.
%\end{itemize}

\subsubsection{Correlation Analysis}
\
Correlation analysis is a statistical method used to evaluate the strength and direction of the relationship between two or more variables. Kendall correlation analysis is a technique suited for determining significant associations between two ranked variables \citep{abdi2007kendall}. A  Kendall correlation coefficient measures the strength of the relationship between two variables, ranging from -1 to 1. Table \ref{ICC} shows a conventional approach to interpreting a correlation coefficient. The correlation coefficient's absolute value closer to 1 indicates a stronger linear relationship between the variables. An absolute value close to 0 suggests negligible correlation. Specifically, a coefficient less than 0.39 indicates a weak correlation; from 0.40 to 0.69, a moderate correlation; from 0.70 to 0.89, a strong correlation; and from 0.90 to 1.00, a very strong correlation \citep{schober2018correlation}. This study used Kendall correlation analysis to understand the extent to which the reasoning complexity of student writing in Chinese correlates with the scoring accuracy of Fine-tuned ChatGPT. 

\begin{table}[h]
\centering
\caption{\textit{A Conventional Approach to Interpreting a Correlation Coefficient \citep{schober2018correlation}}}
\label{ICC}
\begin{tabular}{>{\centering\arraybackslash}p{0.4\linewidth} >{\centering\arraybackslash}p{0.4\linewidth}} 
\hline 
\textbf{Absolute Value of the Observed Correlation Coefficient} & \textbf{Interpretation} \\ 
\hline
0.00-0.09 & Negligible correlation \\ 
0.10-0.39 & Weak correlation \\ 
0.40-0.69 & Moderate correlation \\ 
0.70-0.89 & Strong correlation \\ 
0.90-1.00 & Very strong correlation \\ 
\hline
\end{tabular}
\end{table}

Firstly, we coded the reasoning complexity of students’ responses using the criteria adapted from \cite{Cabello2021elementary,kwon2000linking}. Responses with low reasoning complexity mainly involve generalizing, summarizing, or simple causal reasoning about entities or elements perceptible by the senses. Medium reasoning complexity responses include a few variables elements, or processes beyond the sensory experience, attempting to express causality, nonetheless, not yet at a level that uses the parts of a theory to represent causal processes or ongoing mechanisms. Responses with high reasoning complexity engage with theories or abstract concepts to explain non-visible or non-perceptible elements, processes, or mechanisms, reasoning at this level is at a more sophisticated stage than in the previous levels. It can clarify the covariation relationship among multiple variables, and apply hypothesis deduction, conservation thinking, proportional reasoning, variable control, probability reasoning, and correlation reasoning to construct causal chains or explain running mechanisms.   Table \ref{reasoning_critera} shows the criteria for reasoning complexity and Table \ref{reasoning_complexity} outlines the distribution of students' responses across various reasoning complexity levels in seven items. 

\begin{table}[h]
\centering

\caption{\textit{Criteria for reasoning complexity}}
\label{reasoning_critera}
\begin{tabular}{>{\raggedright\arraybackslash}p{0.14\linewidth}>{\raggedright\arraybackslash}p{0.4\linewidth}>{\raggedright\arraybackslash}p{0.46\linewidth}} 
\hline 
\textbf{Reasoning Complexity}& \textbf{Description}&\textbf{Examplar Responses}\\ 

\hline
High Level & Response includes thinking with non-visible theories or non-perceived elements to explain processes or ongoing mechanisms as the cause of phenomena, using theories, abstract concepts, or models. & As the magnet enters the coil, due to the changing magnetic field inside the coil induces an electric current, heat is generated. According to the equation △Ep = △Ek + Q, as Q is generated, △Ek decreases, resulting in a decrease in △V, which means the magnet's motion is hindered (No. 43).\\ 
Medium Level & Response includes a few variables elements or processes beyond the sensory experience, attempting to express causality, at a level that uses the parts of a theory to represent causal processes or ongoing mechanisms. &The coil will exert resistance on the magnet because of energy conservation (No. 101).\\ 
Low Level & Response mainly involves generalizing, summarizing, or simple causal reasoning about entities or elements within the students’ perception of their senses.& The coil will hinder the fall of the magnet（No.05).\\ 
\hline
\end{tabular}
\end{table}

\begin{table}[h]
\centering
\caption{\textit{Different Levels of Reasoning Complexity}}
\label{reasoning_complexity}
\begin{tabular}{p{0.14\linewidth} p{0.25\linewidth} p{0.25\linewidth} p{0.22\linewidth}} 
\hline 
\textbf{Reasoning Complexity} & \textbf{Low Level} & \textbf{Medium Level} & \textbf{High Level} \\ 
\hline  
\textbf{Item 1} & 9.9\% & 21.9\% & 68.2\% \\ 
\textbf{Item 2} & 43.3\% & 27.4\% & 29.3\% \\ 
\textbf{Item 3} & 35.3\% & 28.1\% & 36.5\% \\ 
\textbf{Item 4} & 3.2\% & 47.8\% & 49.0\% \\ 
\textbf{Item 5} & 10.8\% & 25.3\% & 63.9\% \\ 
\textbf{Item 6} & 22.4\% & 42.7\% & 34.8\% \\ 
\textbf{Item 7} & 7.9\% & 11.6\% & 80.6\% \\ 
\hline 
\end{tabular}

\end{table}
Secondly, we chose the Kendall correlation analysis to assess the correlation between the reasoning complexity of student writing in Chinese and the scoring accuracy of Fine-tuned ChatGPT. We controlled for students' explanation performance during this process, thereby isolating this confounding variable. To control students' performance, we created two performance groups: those scoring 0 were placed in the lower-level group, and those scoring 1 in the higher-level group. Then, we constructed contingency tables to represent the observed frequencies of each combination of reasoning complexity and scoring accuracy categories. Then, we calculated the number of concordant pairs(\textit{n\textsubscript{c}}) and the number of discordant pairs (\textit{n\textsubscript{d}}), the number of samples (\textit{n}), and the minimum value between rows and columns. Following this, we calculated Kendall’s Tau-c using the corresponding formula:

$$\tau_c= \frac{2(n_c - n_d)}{{n^2} \cdot \frac{m-1}{m} }$$
Take Holistic scoring in item 1 as an example. Out of 242 testing samples, 97 scored 0, forming the lower-level group, and 145 scored 1, categorized as the higher-level group. Within the lower-level group, 79 out of 97 samples were with human-machine consistency, 18 samples with inconsistency, 24 out of 97 samples in the lower-level group were coded into low reasoning complexity, 41 samples were coded into medium reasoning complexity,  and 32 samples were coded into high reasoning complexity, as detailed in Table \ref{scoring accuracy*reasoning complexity}. At last, we concluded the Kendall correlation coefficient, τ(97)=-0.32 (\textit{p}$<$0.05), indicating a moderate negative correlation between scoring accuracy and reasoning complexity.
\begin{table}[h]
\centering
\caption{A crosstab example of Scoring Accuracy * Reasoning Complexity in the lower
 level group}
\label{scoring accuracy*reasoning complexity}
\begin{tabular}{>{\centering\arraybackslash}m{2.5cm}>{\centering\arraybackslash}m{1.6cm}>{\centering\arraybackslash}m{1.6cm}>{\centering\arraybackslash}m{1.6cm}>{\centering\arraybackslash}m{1.6cm}} \hline 
   Human-machine &       low reasoning complexity& medium reasoning complexity&high reasoning complexity &total\\ \hline  
 Inconsistency&  0& 7&11 &18\\  
 Consistency& 24& 34& 21&79\\
 Total& 24& 41& 32&97\\\hline 
\end{tabular}

\end{table}

\subsubsection{Qualitative analysis of response characteristics}
\begin{figure}
    \centering
    \includegraphics[width=1\linewidth]{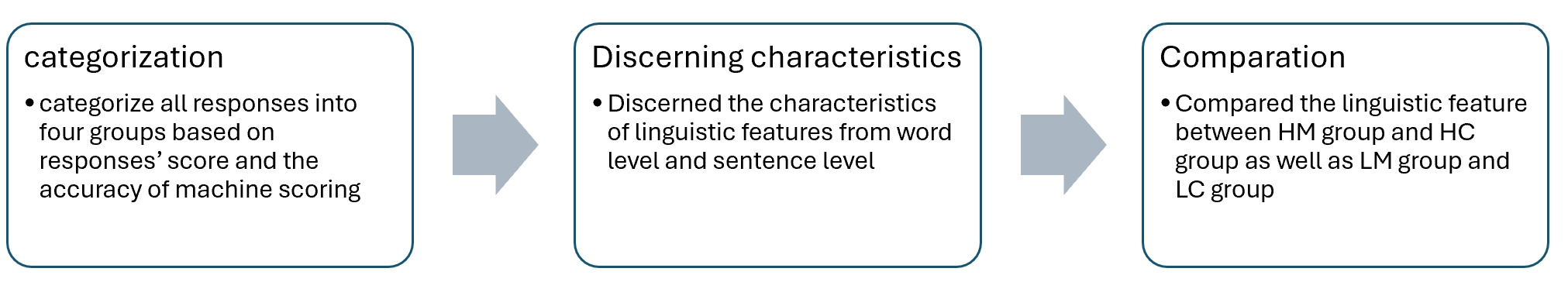}
    \caption{Procedures for the qualitative analysis of responses characteristics}
    \label{fig:Procedures}
\end{figure}

Although the classification process of automatic scoring is usually not transparent, the scoring is based on all information provided. In this study, the input information came from students’ writing responses. Theoretically, all response characteristics could be factors that account for automatic scoring accuracy. Thus, to answer Research Question 3, we employed a qualitative manual matching approach to identify characteristics that might account for scoring differences. 
We analyzed both machine-misscored responses and correctly scored responses, focusing on their distinct languish characteristics following three steps, seeing Figure~\ref{fig:Procedures}.

The first step was categorization, during which researchers segregated the higher-level group and lower-level group into four subgroups based on scoring accuracy: “Higher level and Misscored subgroup (HM subgroup),” “Higher level and Correctly scored subgroup (HC subgroup),” “Lower level and Misscored subgroup (LM subgroup),” and “Lower level and Correctly scored subgroup (LC subgroup).” We then selected all student responses from items where reasoning complexity significantly affected scoring accuracy and categorized them into four subgroups. The second step was discerning characteristics, in which two researchers reviewed all categorized responses to identify linguistic characteristics at both word and sentence levels. According to an analytical framework from \cite{neri2022role}, the researchers determined the linguistic features such as Technical terms, Frequency of words, Polysemous words, Pronouns, Synonyms/Misspellings, and Symbols at the word level, and Syntax, Noun phrases, Verb structures, Number of sentences, Passive constructions, Prepositional phrases, Sentence length, Sentence structure at the sentence level. The third step was a comparison, where researchers compared the linguistic features between the HM and the HC subgroups, as well as between the LM and the LC subgroups. This cross-subgroup comparison aimed to account for discrepancies scored by Fine-tuned ChatGPT. This approach enabled the identification of key linguistic characteristics in student writings that may affect scoring accuracy. Given the diversity of student responses and the large dataset, the matching process was challenging and time-consuming. The findings are also deemed exploratory.
  
\section{Results}

To evaluate the performance of the finetuned ChatGPT on a Chinese dataset, we conducted three sets of experiments to 1) assess the automatic scoring accuracy, 2) examine the correlation between scoring accuracy and reasoning complexity, and 3) analyze the characteristics of students' responses. 

\subsection{Scoring Accuracy of Fine-tuned ChatGPT for Written Responses in Chinese}

\begin{figure}
    \centering
    \includegraphics[width=1\linewidth]{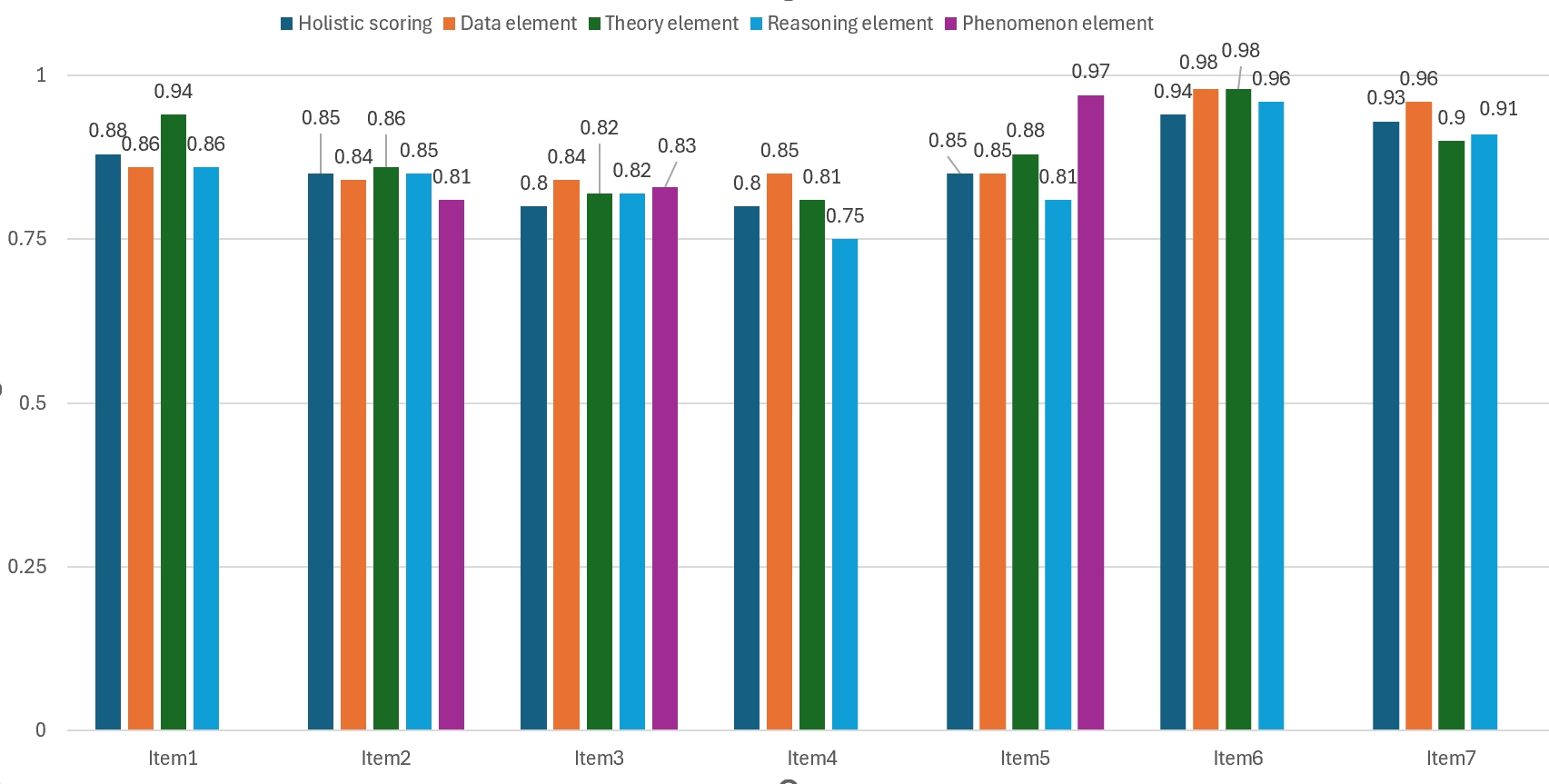}
    \caption{Accuracy results for each item and scoring aspect}
    \label{fig:accuracy_results}
\end{figure}

The accuracy of the finetuned ChatGPT models was examined for several datasets (Item 1 to Item 7; See Figure~\ref{fig:accuracy_results}). The findings indicate that all fine-tuned models achieved scoring accuracy above 0.75 for all scoring categories, though varying by item. Specifically, with scores continuously above 90\%, Items 6 and 7 showed the highest overall accuracy. Item 3 and Item 4, on the other hand, showed worse performance, with all of the scores being below 85\%.

We first examined the model's capacity to accurately assign holistic scores across datasets and found a notable variance, with accuracies ranging from 80\% (Items 3 and 4) to 94\% (Item 6). Further analyses revealed that, for the \textit{Data} element, most items scored in the mid-80s percentage accuracy, indicating a stable performance, except item 6, which had an accuracy of 98\%. In contrast, the \textit{Theory} element yielded an accuracy ranging from 81\% (Item 4) to 98\% (Item 6), implying a relative inconsistency of scoring. Different from the two prior elements, the element \textit{Reasoning} yielded the greatest variation, with Item 4 receiving the lowest accuracy (75\%) and Item 6 achieving the highest (96\%).  Among all the elements, the \textit{Phenomenon} element achieved the highest average accuracy, especially for Item 5, which achieved a remarkable 97\%.

We also found varying accuracy among the four elements for each item. For example, \textit{Theory} element on Item 1 had a high accuracy of 94\%, whilst \textit{Phenomenon} element yielded an accuracy of 86\%. Similarly, Item 4 received 75\% accuracy for \textit{Reasoning} element, which is significantly lower compared to accuracies on other elements. This fluctuation highlights the model's inconsistent reaction to distinct elements of scientific explanations.

The findings show that although the finetuned ChatGPT models exhibit strong performance for certain tasks (e.g., Items 6 and 7), their accuracy varies depending on the scored element and dataset.

\subsection{Association between Reasoning Complexity and ChatGPT Scoring Accuracy }
  
  A Kendall correlation coefficient was calculated to assess the relationship between the reasoning complexity of students’ responses and the scoring accuracy for each scoring element across seven items. Considering that students' explanation performance is a potential confounding variable affecting both reasoning complexity and scoring accuracy, we controlled students'  performance to more accurately isolate the correlation. Specifically, we grouped students into lower- and higher-level groups based on their overall performance and then analyzed the relationships by students' performance level.
 
\subsubsection{The lower-level group }

The lower-level group received 0 for their responses, the correlation between reasoning complexity and scoring accuracy for the lower-level group responses is shown in Table \ref{Lower}, we calculated a total of 30 Kendall correlations to explore the relationships between reasoning complexity and scoring accuracy across different scoring elements for seven items. Among these correlations, 29 correlations were negative, with 28 being statistically significant (\textit{p }$<$0.05 or \textit{p} $<$0.1).  The average correlation coefficient across all items and scores was -0.31 (SD = 0.17). These results indicate that for lower-level performing students, the higher the complexity of reasoning in their explanations, the lower the scoring accuracy yielded by ChatGPT models. 

%Particularly, eight instances exhibited negative moderate correlations(-0.69$<$τ$<$-0.40), indicating a substantial relationship where higher reasoning complexity is associated with lower scoring accuracy. Nineteen instances showed weak correlations (-0.39$<$ τ$<$-0.10), suggesting a  noticeable but not particularly strong connection between reasoning complexity and scoring accuracy, where increased reasoning complexity tends to slightly lower scoring accuracy. One instance presented a negligible correlation (τ=-0.09, P$<$0.05), indicating very little relationship between reasoning complexity and scoring accuracy in this particular case.

  \begin{table}[h]
\centering
\caption{Correlation between reasoning complexity and scoring accuracy for lower-level performance responses}
\label{Lower}
\begin{tabular}{>{\centering\arraybackslash}m{1.6cm}>{\centering\arraybackslash}m{1.3cm}>{\centering\arraybackslash}m{1.3cm}>{\centering\arraybackslash}m{1.3cm}>{\centering\arraybackslash}m{1.3cm}>{\centering\arraybackslash}m{1.6cm}>{\centering\arraybackslash}m{1.2cm}}\hline
 & \multicolumn{6}{c}{Scoring Accuracy}\\\hline 
    Reasoning Complexity& Holistic scoring& Data element& Theory element& Reasoning element& Phenomenon element& Mean   \\ \hline  
 Item1& -0.32**& 0.15& -0.09*& -0.19*& /& -0.11*\\   
 Item2& -0.35**& -0.25**& -0.25**& -0.37**& -0.35**& -0.31**\\   
Item3& -0.62**& -0.19**& -0.34**& -0.45**& -0.52**& -0.42**\\   
Item4& -0.27**& -0.20**& -0.24**& -0.31**& /& -0.26**\\ 
 Item5& -0.43**& -0.34**& -0.12**& -0.36**& -0.61**&-0.38**\\  
Item6& -0.09**& -0.02& /& -0.27**& /&-0.13\\  
 Item7& -0.40**& -0.55**& -0.53**& -0.25**& /&-0.43**\\  
 Mean
& -0.35**& -0.20& -0.26**& -0.37**& -0.50**&-0.34**\\ \hline 
\end{tabular}
\begin{minipage}{0.95\textwidth} \footnotesize 
\textbf{Notes} 
\begin{itemize}
\item   * \textit{p} $<$ .1, ** \textit{p} $<$ .05 
\item the scoring accuracy of Theory element in Item 6 is 100\% 
\item   Phenomenon element does not apply to Item 1, Item 4, Item 6, and Item 7, since the phenomenon was given in these items.
\end{itemize} 
\end{minipage}
\end{table}
\subsubsection{The higher-level group }
 The higher-level group received 1 for their responses, we conducted the same total of 30 Kendall correlation analyses to assess the relationship between reasoning complexity and scoring accuracy for the higher-level group responses, Table \ref{Higher} shows the results. Among these, 26 out of 30 correlations were positive, with 10 being significant (\textit{p} $<$0.05 or \textit{p} $<$0.1).  The average value of these significant correlation coefficients was 0.19 (SD =0.11 ). These results indicate that for higher-level students, the higher the reasoning complexity is, the higher of scoring accuracy yielded by ChatGPT.
 
% Specifically, eight instances exhibited weak correlations (0.10$<$ τ$<$0.39), suggesting a noticeable but not particularly strong connection between reasoning complexity and scoring accuracy, where increased reasoning complexity tends to slightly higher scoring accuracy. One instance showed a negligible correlation (τ=0.09, P$<$0.05), indicating a very small but significant relationship between reasoning complexity and scoring accuracy in this particular case.

\begin{table}
\centering
\caption{Correlation between reasoning complexity and grading accuracy for higher-level responses}
\label{Higher}
\begin{tabular}{>{\centering\arraybackslash}m{0.15\linewidth}>{\centering\arraybackslash}m{1cm}>{\centering\arraybackslash}m{1.3cm}>{\centering\arraybackslash}m{1.3cm}>{\centering\arraybackslash}m{1.3cm}>{\centering\arraybackslash}m{1.5cm}>{\centering\arraybackslash}m{1cm}}\hline
 & \multicolumn{6}{c}{Scoring Accuracy }\\\hline 
   Reasoning complexity &  Holistic scoring&  
Data element& Theory element& Reasoning element& Phenomenon element& Mean \\ \hline  
Item1& 0.03& 0.07& -0.01*& 0.02& /& 0.01\\   
Item2& -0.02& -0.02& 0.17& 0.07& 0.15*& 0.07\\  
Item3& 0.14& 0.10& 0.37**& 0.03& 0.00& 0.13\\   
Item4& 0.09**& 0.31**& 0.00& 0.30**& /& 0.17\\ 
 Item5& 0.01& 0.09& 0.05& 0.16**& 0.03&0.07\\  
 Item6& 0.21*& 0.01& -0.06& 0.17**& /&0.08\\  
 Item7& 0.05& /& 0.10**& 0.06& /&0.07*\\  
 Mean
& 0.07& 0.09& 0.09& 0.17& 0.06&0.09\\ \hline 
\end{tabular}
\begin{minipage}{0.95\textwidth} \footnotesize 
\textbf{\textit{Note.}}
\begin{itemize}
\item   \textit{*p }$<$ .1, \textit{**p} $<$ .05 
\item The scoring accuracy for the \textit{Data} element in Item 7 is 100\% 
\item   \textit{Phenomenon} element does not apply to Item 1, Item 4, Item 6, and Item 7, since the phenomenon was not required for students in these items.
\end{itemize} 
\end{minipage}

\end{table}

\subsection{The linguistic features of students' responses}

In this section, we present student’s writing responses’ linguistic features both at the word and sentence levels, the features may account for the machine mislabels. Since the most basic written unit in Chinese is a character rather than a word, the frequency calculations in this study are based on the frequency of Chinese characters instead of the frequency of words.  

\subsubsection{The characteristics of the lower-level group }

Overall, responses from the lower-level group received 0 for their responses and demonstrated a low frequency of misspellings, polysemous words, synonyms, and pronouns, tend to use academic terms, and describe phenomena directly at the word level, rarely use passive constructions at the sentence level. Additionally, we also obtain some distinct patterns between the correctly scored subgroup and the misscored subgroup.

\textbf{\textit{The lower level and correctly scored subgroup. }}At the word level, the written responses from the Lower level and Correctly scored subgroup (LC subgroup) used fewer technical terms, with a higher proportion of everyday language, and averaged only 20 Chinese characters, significantly shorter than the 40 Chinese characters found in the Lower level and Misscored subgroup (LM subgroup).  At the sentence level, the LC subgroup exhibited several distinct characteristics compared to the LM subgroup:(1) they contain fewer and shorter sentences, typically with simple structures like subject-verb-object; (2) the syntax was relatively straightforward, using more simple and compound sentences; (3) noun and verb phrases were more straightforward, and prepositional phrases were less frequent, making the descriptions more comprehensible. 

Take a typical LC subgroup response in item 7 as an example. The response was \textit{“}\textit{There are gaps between molecules that allow for transmission (No. 194).”} This response only describes one phenomenon and fails to incorporate theories related to molecular thermal motion. Therefore, it receives a human score of 0 in the \textit{Theory} element. It acknowledges the interaction between molecules at the microscopic level but lacks reasoning chains to show why the gaps between molecules lead to transmission, resulting in a lower reasoning complexity. 
 
Moreover, this response is short, containing only ten Chinese characters, and uses everyday terms like \textit{“gaps”} and \textit{“transmission”.} Its structure is simple, with just one subject-verb-object sentence, and lacks complex clauses or modifiers. The phrases, such as “\textit{gaps between molecules}” were straightforward and clear, without complex technical terms or multilayered modifications. This direct and easily understandable structure and terms aided in conveying the information more clearly. Due to the simplicity and directness, the machine scoring system can more easily parse this type of response. The simple sentence structure reduces the likelihood of parsing errors and makes information extraction straightforward and efficient. This concise and clear expression style helps the machine algorithm accurately evaluate the student's response, allowing the machine scoring to achieve high accuracy in this context, as the scoring criteria align more easily with the straightforward information provided by the student.

\textbf{\textit{The lower level and misscored subgroup}.} Unlike the LC subgroup, the written responses by the Lower level and Misscored subgroup (LM subgroup) tend to be longer, averaging about 40 Chinese characters. These responses frequently use technical terms and professional vocabulary but not always accurately, especially when describing physical processes. At the sentence level, LM subgroup responses typically contain multiple sentences, providing detailed explanations. Their sentence structures are complex, with varying lengths and multiple modifying elements. They frequently used complex syntactic structures, including multiple clauses and relative clauses. Additionally, these responses feature complex and technical noun phrases and verb phrases, and they use prepositional phrases frequently, particularly when describing physical locations and states. These linguistic characteristics often result in higher reasoning complexity. However, they may also complicate the understanding and parsing of these responses, potentially challenging the scoring accuracy of ChatGPT.

For instance, the response to item 7 is “\textit{The principle of diffusion is that molecules are constantly in random motion. The higher the temperature, the greater the internal energy, the internal energy is converted into kinetic energy and released, causing vigorous molecular motion, increased gaps, reduced forces, and greater diffusion (No. 59)}.” The response displays a misconception-- “\textit{internal energy is converted into kinetic energy}”, which led to a score of 0 in the \textit{Theory} element. However, it surpasses sensory experience by using a microscopic molecular thermal motion model and forming an evidence-to-theory-to-phenomenon reasoning chain, indicating a high reasoning complexity level. 

This response exceeds 60 Chinese characters and uses numerous technical terms such as “\textit{kinetic energy}” “\textit{internal energy}” and “\textit{force}”. Regarding sentence structure, the response contains a long sentence segmented by commas into various clauses and phrases and contains information equivalent to multiple shorter sentences. Each clause carries a complete idea. For instance, \textit{“The higher the temperature, the greater the internal energy, the internal energy is converted into kinetic energy and released”} shows a progressive structure, adding complexity and information. Besides, the syntax is relatively complex, employing causal clauses and explanatory phrases. For example, \textit{“The principle of diffusion is”} introduces a definition followed by several clauses and phrases attempting to explain the mechanism of diffusion. This use of multiple causal relationships and parallel structures enhances the complexity of the sentence. It also employs multiple complex noun and verb phrases, such as \textit{“molecules are constantly in random motion”} and \textit{“internal energy is converted into kinetic energy and released,”} which are both scientifically rigorous and descriptive of specific physical processes, demonstrating high technicality and professionalism in language use. 

The sentence's structure is intricate, involving several physical processes and transformations, potentially making it difficult for the ChatGPT to accurately capture the logical relationships and scientific concepts within each clause. This complexity necessitates a scoring algorithm capable of high semantic understanding and precise handling of scientific terminology and logical structures, which may result in less accurate scoring and lead to a high risk of misscored, where students with lower levels of performance may receive higher scores.
 
\subsubsection{The characteristics of the Higher-level group }

Compared to lower-performing responses, higher-performing ones included scientific terms more frequently. They used polysemous words, pronouns, and synonyms less often. Additionally, distinctive patterns emerged between the Higher level and Correctly scored (HC) subgroup and the Higher level and Misscored (HM) subgroup.

\textbf{\textit{The higher level and correctly scored subgroup.}} At the word level, responses from the HC subgroup averaged about 46 Chinese characters, which had 20 more characters compared with the HM subgroup. Additionally, the HC subgroup used mathematical symbols or formulas more frequently than the HM subgroup. The longer responses and the use of symbols/formulas offered more detailed and precise explanations. At the sentence level, the HC subgroup's responses exhibited several features distinct from those of the HM subgroup: (1) most responses of the HC subgroup included multiple sentences with complex syntactic structures, incorporating multiple clauses and modifiers; (2) They often employed numerous noun phrases to explain phenomena, such as ``\textit{rate of change of magnetic flux}"; (3) Verb structures were more complex and varied, such as “\textit{partially converted the mechanical energy of the magnet into thermal energy}”; (4) Each student response contained a higher number of sentences, typically ranging from 5 to 10; (5) Prepositional phrases were also more prevalent, such as ``\textit{entering the coil}" and ``\textit{during the free fall of the magnet}" clearly indicating action states.  

For instance, the response to item 3, ``\textit{As the magnet enters the coil, due to the changing magnetic field inside the coil induces an electric current, heat is generated. According to the equation }\textit{△Ep = △Ek + Q, as Q is generated, △Ek decreases, resulting in a decrease in △V, which means the magnet's motion is hindered (No. 43)}," accurately applies the principle of energy conservation and correctly uses the equation, thus receiving a score of 1 on the \textit{Theory} dimension. It effectively employs non-visible theories and equations to link multiple related variables, explaining the mechanism of the magnet's fall. Therefore, it demonstrates a higher reasoning complexity level. 

This response exceeded 70 Chinese characters, used numerous technical terms, such as \textit{``electric current"} and\textit{ ``magnetic field"},  and employed symbols like \textit{``Q"}, \textit{``△Ek"}, and \textit{``△V"}, as well as related equations to precisely describe the physical phenomena and processes, reducing ambiguity. It also includes multiple prepositional phrases, such as \textit{``due to the change in the magnetic field inside the coil”} and \textit{``as Q is generated,}” providing temporal and causal background information for a more complete description. Although it is a long sentence, it uses commas to separate different clauses and phrases, maintaining a strict logical progression. Starting from the fact that ``\textit{the magnet enters the coil}," it connects through the theory ``\textit{△Ep = △Ek + Q,}" and finally concludes with ``\textit{the magnet's motion is hindered}," thoroughly explaining the phenomenon of the magnet’s fall. This complete and precise expression style reduces the likelihood of parsing errors and helps the machine algorithm accurately evaluate the student's response, allowing the machine scoring to achieve high accuracy in this context, as the machine has more accurate information provided by the student to compare with the scoring criteria.

\textbf{\textit{The Higher level and Misscored subgroup.}} Responses from the HM subgroup averaged about 26 Chinese characters. These responses seldom use mathematical symbols and formulas, relying mainly on textual explanations. The syntax was not complex, consisting mostly of simple and short sentences, which made them easy to understand. The noun phrases used were limited and simple, such as ``\textit{change in a magnetic field}" and ``\textit{The work done by the pump on the gas}". The verb structures are relatively straightforward, like ``\textit{induces an electric current}" and ``\textit{generate heat}". Compared with the HC subgroup, each response in the HM subgroup contained fewer sentences, typically ranging from 1 to 3, and prepositional phrases were fewer and simpler, such as ``\textit{doing work on the gas}" resulting in an unclear representation of physical states. 

Taking the HM subgroup response to item 3 as an example, the response, ``\textit{The coil will exert resistance on the magnet because of energy conservation (No. 101),}" employs the principle of energy conservation to provide a concise explanation of the phenomenon, earning a score of 1 on the theory element. However, it simply links energy conservation with the conclusion of resistance, offering some causal analyses but failing to explain the mechanism of the magnet’s fall. Therefore, it demonstrates a medium reasoning complexity. 

This response contains only 19 Chinese characters and uses mostly non-technical terms, such as \textit{``resistance}" instead of precise terms like ``\textit{repulsive force}" or ``\textit{repulsion}," which may lead to ambiguity in model interpretation. The sentences do not include mathematical symbols or formulas, and the syntax is simple, consisting of only a single simple sentence. It lacks prepositional phrases to describe the state of physical phenomena, resulting in a lack of detailed contextual explanation and specific physical mechanism inference.  These linguistic features will lead to a risk of being misscored, where students with higher performance levels may receive lower scores.

\section{Discussions}

\subsection{Fine-tuned ChatGPT Can Automatic Score Student Written Explanations in Chinese, though Less Accurate than Those Written in English}
  
Automatic scoring has long been regarded as essential for effective assessment practices in education. Various automatic scoring algorithms and strategies have been employed to achieve high scoring accuracy \citep{fiacco-etal-2022-toward,gombert2023coding,zhai2020applying} with varying degrees of success \citep{zhai2021meta}. Prior research has demonstrated the potential of LLMs in automatically scoring student written responses \citep{liu2023context,riordan2020empirical}. In our study, we examined the accuracy of Fine-tuned ChatGPT in scoring scientific explanations in Chinese. The results revealed high agreements between human and machine scoring, consistent with findings from English contexts, such as \cite{latif2024fine}. Our study offers evidence for its usability across different linguistic and cultural environments, highlighting its potential for broader applications. We also found that although the finetuned ChatGPT models exhibit strong performance for certain tasks, their accuracy varies depending on the scored element and dataset. These differences may be due to the scoring criteria' intricacy, the student answers' characteristics in each dataset, or the model's intrinsic incapacity to comprehend and assess particular kinds of scientific explanations. More research into these areas might yield more insightful conclusions and possibly direct advancements in techniques for fine-tuning and training models. 

Additionally, we observed that the accuracy in a Chinese context is not as high as in English contexts \citep{latif2024fine}, despite using similar scoring rubrics and sample sizes. This discrepancy is likely due to the unique linguistic characteristics of Chinese, such as its implicit, circular, visual, and synthetic reasoning styles, which are consistent with findings from \citep{Wang2021automated}. These identified characteristics pose challenges for current language models \citep{Wu2023GPTOverview, John2024GPTreasoning}. This study also found noticeable differences in accuracy among various scores for each item, even when inputting the same written responses. However, unlike previous research by \cite{Wang2021automated}, we did not indicate higher accuracy in analytic scoring compared to holistic methods. We believe this is due to the clarity and specificity of our scoring rubrics, which minimized differences in scoring precision. This finding suggests that the primary determinant of scoring accuracy by machines is the quality of human judgment used in model training and rubric design, rather than the type of scoring rubric employed \citep{beggrow2014assessing}.Our results demonstrate significant potential for practical application in classroom teaching. For teachers, the model can assist in automatically scoring students' written responses in Chinese and providing immediate feedback, thereby reducing grading workload and enabling more efficient assessment practices. This immediate feedback also helps teachers make informed instructional decisions, such as identifying common misconceptions and adjusting teaching strategies to meet students' needs better\citep{adiguzel2023revolutionizing}. For students, timely feedback on their written responses can support their personalized learning by reinforcing correct concepts and promptly addressing errors\citep{limna2022review}. These future applications highlight the practical value of fine-tuned language models in enhancing both teaching effectiveness and student engagement in Chinese language classrooms.

\subsection{The Scoring Accuracy is Associated with Reasoning Complexity in Writing}

Further analyses revealed a significant correlation between the complexity of reasoning of student responses and scoring accuracy.  In the lower-level group, we observed that as the complexity of reasoning in student responses increased, the scoring accuracy decreased. This negative correlation suggests that implicit and complex reasoning expressions pose significant challenges for machines to score accurately\citep{John2024GPTreasoning}. Conversely, within the higher-level group, a counterintuitive trend was noted, which shows that a decrease in reasoning complexity for some items coincided with reduced scoring accuracy.

The results reflect the differing limitations of using fine-tuned ChatGPT for automatic scoring when evaluating responses involving different levels of student performance. For the low-level performing student responses, the LLMs tend to assign a higher score than expected to their responses, once they include complexity reasoning, such as thinking with multiple reasoning or reasoning chains and using non-visible theories or non-perceived elements to explain processes or ongoing mechanisms as the cause of phenomena, even if the reasoning process is entirely incorrect. Similarly, \cite{haudek2023examining} has found that complex and diverse evaluation tasks, which often mean low-level and diverse student responses, can lead to lower model performance. In contrast, for the high-level performing student responses, the LLMs tend to assign lower scores than expected if these responses use generalizing, summarizing, or simple causal reasoning, and adapt a few variable elements or processes beyond sensory experience instead of theories, abstract concepts, or models,  attempting to express causality. These limitations highlight the challenges of accurately scoring complex reasoning and abstract thinking in student responses using fine-tuned ChatGPT.

 \subsection{The Languish Characteristics and Scoring Accuracy}

Chinese characters inherently exhibit linguistic structural diversity\citep{li2019word}, and implicit expression characteristics, which have a significant effect on the construction of students' scientific reasoning\citep{Williamsf2010Chinese}. Previous research has demonstrated that language features, such as the length of responses, linguistic comprehensiveness, and the use of specific formulas and terminology affected the accuracy of automated scoring\citep{gombert2023coding,lockwood2015handbook}. However, findings on how these characteristics impact scoring accuracy have been mixed. For instance, \cite{Wang2021automated} observed that shorter Chinese writings tend to be scored accurately, due to more precise meaning, while \cite{Nehm2012Transforming} found that scoring models were least accurate for the shortest response responses. 

In our study, we observed both types of outcomes. Through qualitative analyses of linguistic features, we identified unique patterns in the language styles of high-performing and low-performing groups. These patterns may partially explain the contradictory results. In the LM subgroup, responses frequently exhibited inaccurate use of terminology, complex sentence structures, and varying sentence lengths, with a high level of reasoning complexity. However, as syntactic language complexity increased, the difficulty of parsing these responses also increased, making it harder for the machine to understand the students’ intended meaning accurately. Consequently, ChatGPT demonstrated a higher likelihood of generating inaccurate scoring \citep{dhar2023we}. Conversely, the LC subgroup provided shorter, clearer, and more focused responses with lower reasoning complexity, making them easier to parse. This finding is consistent with the result from \cite{Wang2021automated}, and  further explains why lower reasoning complexity corresponds to higher coring accuracy.

Conversely, the high-performing group exhibited contrasting trends. The HM subgroup often provided concise answers that, despite being easier to parse, often lacked linguistic comprehensiveness, resulting in lower reasoning complexity levels and frequent mislabeling by the machine. In contrast, the HC subgroup employed complex sentence structures with strong internal logic and longer sentences, including detailed explanations and information that enabled more accurate judgments by the machine, consistent with the results from \cite{Nehm2012Transforming}. Additionally, the HC subgroup’s use of rich terminology and mathematical symbols/formulas demonstrates a precise expression of scientific explanations. Automated scoring systems favor such detailed and structured responses \citep{lockwood2015handbook}, reinforcing why higher reasoning complexity corresponds to higher scoring accuracy in the high-performing group. These insights underscore the need for automated scoring systems to adapt to the linguistic complexity and reasoning levels presented in student responses, promoting a deeper understanding of how AI can better support diverse educational assessments.

\section{Conclusions and Limitation}
Our study investigated the potential of fine-tuned ChatGPT for automatic scoring in non-English contexts, specifically within Chinese educational settings. Through domain-specific adaptation, the fine-tuned scoring models effectively capture the complexities and nuances unique to educational datasets \citep{latif2024fine}. Our findings indicate that fine-tuning models can accurately score students’ Chinese written responses, underscoring the technology’s potential as a cutting-edge tool for educational applications such as automatic scoring in multicultural settings, particularly in Chinese.  Furthermore, Our findings reveal a nuanced relationship between reasoning complexity and scoring accuracy: lower-level responses show a significant negative correlation, while higher-level responses exhibit a positive correlation. This seemingly contradictory result can be attributed to differences in the linguistic features of the two groups of students. The comprehensiveness and accuracy of student responses are often in tension with the simplicity and clarity of language structure. When simplicity and clarity are prioritized, simpler and shorter responses are scored more accurately, as demonstrated by \cite{Wang2021automated}. Conversely, when comprehensiveness is prioritized, responses may be scored less accurately, as indicated by \cite{Nehm2012Transforming}. These insights emphasize the potential biases in scoring the highly abstract or minimally complex responses as well as highlight the importance of tailoring ChatGPT to specific educational contexts to optimize scoring accuracy. 

Despite these promising results, our study has limitations. First, the exploratory nature of our analyses of the associations between linguistic features with scoring accuracy underscores the need for a more structured quantitative framework. Second, while we focused on the ChatGPT model, predicting similar outcomes with other LLMs remains hypothetical and requires empirical validation. Future research should extend these findings across various AI models and explore different linguistic and cultural settings to validate and expand our understanding. Lastly, as we integrate AI tools like ChatGPT into educational practices, it is imperative to address potential ethical concerns, including bias, privacy, and data security \cite{zhai2023ai}. Ensuring that these technologies are used responsibly and effectively in educational settings is essential to uphold ethical standards and enhance their utility.

\nolinenumbers

\end{CJK*}

\bibliography{sn-bibliography}% common bib file

\begin{thebibliography}{}
\renewcommand{\doi}[1]{\url{https://doi.org/#1}}
\bibcommenthead

\bibitem [\protect \citeauthoryear {%
Abdi%
}{%
Abdi%
}{%
{\protect \APACyear {2007}}%
}]{%
abdi2007kendall}
\APACinsertmetastar {%
abdi2007kendall}%
\begin{APACrefauthors}%
Abdi, H.%
\end{APACrefauthors}%
\unskip\
\newblock
\APACrefYearMonthDay{2007}{}{}.
\newblock
{\BBOQ}\APACrefatitle {The Kendall rank correlation coefficient} {The kendall rank correlation coefficient}.{\BBCQ}
\newblock
\APACjournalVolNumPages{Encyclopedia of Measurement and Statistics. Sage, Thousand Oaks, CA}{}{}{508--510,}
\newblock

\newblock

\PrintBackRefs{\CurrentBib}

\bibitem [\protect \citeauthoryear {%
Adiguzel%
, Kaya%
\BCBL {}\ \BBA {} Cansu%
}{%
Adiguzel%
\ \protect \BOthers {.}}{%
{\protect \APACyear {2023}}%
}]{%
adiguzel2023revolutionizing}
\APACinsertmetastar {%
adiguzel2023revolutionizing}%
\begin{APACrefauthors}%
Adiguzel, T.%
, Kaya, M.H.%
\BCBL {} Cansu, F.K.%
\end{APACrefauthors}%
\unskip\
\newblock
\APACrefYearMonthDay{2023}{}{}.
\newblock
{\BBOQ}\APACrefatitle {Revolutionizing education with AI: Exploring the transformative potential of ChatGPT} {Revolutionizing education with ai: Exploring the transformative potential of chatgpt}.{\BBCQ}
\newblock
\APACjournalVolNumPages{Contemporary Educational Technology}{15}{3}{ep429,}
\newblock

\newblock

\PrintBackRefs{\CurrentBib}

\bibitem [\protect \citeauthoryear {%
Ariely%
, Nazaretsky%
\BCBL {}\ \BBA {} Alexandron%
}{%
Ariely%
\ \protect \BOthers {.}}{%
{\protect \APACyear {2023}}%
}]{%
ariely2023machine}
\APACinsertmetastar {%
ariely2023machine}%
\begin{APACrefauthors}%
Ariely, M.%
, Nazaretsky, T.%
\BCBL {} Alexandron, G.%
\end{APACrefauthors}%
\unskip\
\newblock
\APACrefYearMonthDay{2023}{}{}.
\newblock
{\BBOQ}\APACrefatitle {Machine learning and Hebrew NLP for automated assessment of open-ended questions in biology} {Machine learning and hebrew nlp for automated assessment of open-ended questions in biology}.{\BBCQ}
\newblock
\APACjournalVolNumPages{International journal of artificial intelligence in education}{33}{1}{1--34,}
\newblock

\newblock

\PrintBackRefs{\CurrentBib}

\bibitem [\protect \citeauthoryear {%
Beaty%
\ \BBA {} Johnson%
}{%
Beaty%
\ \BBA {} Johnson%
}{%
{\protect \APACyear {2021}}%
}]{%
beaty2021automating}
\APACinsertmetastar {%
beaty2021automating}%
\begin{APACrefauthors}%
Beaty, R.E.%
\BCBT {}\ \BBA {} Johnson, D.R.%
\end{APACrefauthors}%
\unskip\
\newblock
\APACrefYearMonthDay{2021}{}{}.
\newblock
{\BBOQ}\APACrefatitle {Automating creativity assessment with SemDis: An open platform for computing semantic distance} {Automating creativity assessment with semdis: An open platform for computing semantic distance}.{\BBCQ}
\newblock
\APACjournalVolNumPages{Behavior research methods}{53}{2}{757--780,}
\newblock

\newblock

\PrintBackRefs{\CurrentBib}

\bibitem [\protect \citeauthoryear {%
Beggrow%
, Ha%
, Nehm%
, Pearl%
\BCBL {}\ \BBA {} Boone%
}{%
Beggrow%
\ \protect \BOthers {.}}{%
{\protect \APACyear {2014}}%
}]{%
beggrow2014assessing}
\APACinsertmetastar {%
beggrow2014assessing}%
\begin{APACrefauthors}%
Beggrow, E.P.%
, Ha, M.%
, Nehm, R.H.%
, Pearl, D.%
\BCBL {} Boone, W.J.%
\end{APACrefauthors}%
\unskip\
\newblock
\APACrefYearMonthDay{2014}{}{}.
\newblock
{\BBOQ}\APACrefatitle {Assessing scientific practices using machine-learning methods: How closely do they match clinical interview performance?} {Assessing scientific practices using machine-learning methods: How closely do they match clinical interview performance?}{\BBCQ}
\newblock
\APACjournalVolNumPages{Journal of Science education and Technology}{23}{}{160--182,}
\newblock

\newblock

\PrintBackRefs{\CurrentBib}

\bibitem [\protect \citeauthoryear {%
Bjerknes%
, Wilhelmsen%
\BCBL {}\ \BBA {} Foyn-Bruun%
}{%
Bjerknes%
\ \protect \BOthers {.}}{%
{\protect \APACyear {2024}}%
}]{%
Bjerknes2024curiosity}
\APACinsertmetastar {%
Bjerknes2024curiosity}%
\begin{APACrefauthors}%
Bjerknes, A\BHBI L.%
, Wilhelmsen, T.%
\BCBL {} Foyn-Bruun, E.%
\end{APACrefauthors}%
\unskip\
\newblock
\APACrefYearMonthDay{2024}{}{}.
\newblock
{\BBOQ}\APACrefatitle {A Systematic Review of Curiosity and Wonder in Natural Science and Early Childhood Education Research} {A systematic review of curiosity and wonder in natural science and early childhood education research}.{\BBCQ}
\newblock
\APACjournalVolNumPages{Journal of Research in Childhood Education}{38}{1}{50--65,}
\newblock

\newblock

\PrintBackRefs{\CurrentBib}

\bibitem [\protect \citeauthoryear {%
Cabello%
, Moreira%
\BCBL {}\ \BBA {} Morales%
}{%
Cabello%
\ \protect \BOthers {.}}{%
{\protect \APACyear {2021}}%
}]{%
Cabello2021elementary}
\APACinsertmetastar {%
Cabello2021elementary}%
\begin{APACrefauthors}%
Cabello, V.%
, Moreira, P.%
\BCBL {} Morales, P.G.%
\end{APACrefauthors}%
\unskip\
\newblock
\APACrefYearMonthDay{2021}{}{}.
\newblock
{\BBOQ}\APACrefatitle {Elementary students’ reasoning in drawn explanations based on a scientific theory} {Elementary students’ reasoning in drawn explanations based on a scientific theory}.{\BBCQ}
\newblock
\APACjournalVolNumPages{Education Sciences}{11}{10}{581,}
\newblock

\newblock

\PrintBackRefs{\CurrentBib}

\bibitem [\protect \citeauthoryear {%
Chen%
\ \BBA {} Zhang%
}{%
Chen%
\ \BBA {} Zhang%
}{%
{\protect \APACyear {2021}}%
}]{%
chen2021influence}
\APACinsertmetastar {%
chen2021influence}%
\begin{APACrefauthors}%
Chen, H.%
\BCBT {}\ \BBA {} Zhang, Y.%
\end{APACrefauthors}%
\unskip\
\newblock
\APACrefYearMonthDay{2021}{}{}.
\newblock
{\BBOQ}\APACrefatitle {The Influence of Cultural Differences between Chinese and English on Translation [J]} {The influence of cultural differences between chinese and english on translation [j]}.{\BBCQ}
\newblock
\APACjournalVolNumPages{Cross Current Int J Peer Reviewed J Human Soc Sci}{7}{4}{67--72,}
\newblock

\newblock

\PrintBackRefs{\CurrentBib}

\bibitem [\protect \citeauthoryear {%
De~Andrade%
, Freire%
\BCBL {}\ \BBA {} Baptista%
}{%
De~Andrade%
\ \protect \BOthers {.}}{%
{\protect \APACyear {2019}}%
}]{%
de2019constructing}
\APACinsertmetastar {%
de2019constructing}%
\begin{APACrefauthors}%
De~Andrade, V.%
, Freire, S.%
\BCBL {} Baptista, M.%
\end{APACrefauthors}%
\unskip\
\newblock
\APACrefYearMonthDay{2019}{}{}.
\newblock
{\BBOQ}\APACrefatitle {Constructing scientific explanations: A system of analysis for students’ explanations} {Constructing scientific explanations: A system of analysis for students’ explanations}.{\BBCQ}
\newblock
\APACjournalVolNumPages{Research in Science Education}{49}{}{787--807,}
\newblock

\newblock

\PrintBackRefs{\CurrentBib}

\bibitem [\protect \citeauthoryear {%
Dhar%
\ \BBA {} Bose%
}{%
Dhar%
\ \BBA {} Bose%
}{%
{\protect \APACyear {2023}}%
}]{%
dhar2023we}
\APACinsertmetastar {%
dhar2023we}%
\begin{APACrefauthors}%
Dhar, S.%
\BCBT {}\ \BBA {} Bose, I.%
\end{APACrefauthors}%
\unskip\
\newblock
\APACrefYearMonthDay{2023}{}{}.
\newblock
{\BBOQ}\APACrefatitle {Are We Nearing Singularity? A Study of Language Capabilities of ChatGPT} {Are we nearing singularity? a study of language capabilities of chatgpt}.{\BBCQ}
\newblock
 \APACrefbtitle {Analytics Global Conference} {Analytics global conference}\ (\BPGS\ 125--135).
\PrintBackRefs{\CurrentBib}

\bibitem [\protect \citeauthoryear {%
Dougrez-Lewis%
, Akhter%
, He%
\BCBL {}\ \BBA {} Liakata%
}{%
Dougrez-Lewis%
\ \protect \BOthers {.}}{%
{\protect \APACyear {2024}}%
}]{%
John2024GPTreasoning}
\APACinsertmetastar {%
John2024GPTreasoning}%
\begin{APACrefauthors}%
Dougrez-Lewis, J.%
, Akhter, M.E.%
, He, Y.%
\BCBL {} Liakata, M.%
\end{APACrefauthors}%
\unskip\
\newblock
\APACrefYearMonthDay{2024}{}{}.
\newblock
{\BBOQ}\APACrefatitle {Assessing the Reasoning Abilities of ChatGPT in the Context of Claim Verification} {Assessing the reasoning abilities of chatgpt in the context of claim verification}.{\BBCQ}
\newblock
\APACjournalVolNumPages{arXiv preprint arXiv:2402.10735v2}{}{}{,}
\newblock

\newblock

\PrintBackRefs{\CurrentBib}

\bibitem [\protect \citeauthoryear {%
Driver%
, Newton%
\BCBL {}\ \BBA {} Osborne%
}{%
Driver%
\ \protect \BOthers {.}}{%
{\protect \APACyear {2000}}%
}]{%
driver2000establishing}
\APACinsertmetastar {%
driver2000establishing}%
\begin{APACrefauthors}%
Driver, R.%
, Newton, P.%
\BCBL {} Osborne, J.%
\end{APACrefauthors}%
\unskip\
\newblock
\APACrefYearMonthDay{2000}{}{}.
\newblock
{\BBOQ}\APACrefatitle {Establishing the norms of scientific argumentation in classrooms} {Establishing the norms of scientific argumentation in classrooms}.{\BBCQ}
\newblock
\APACjournalVolNumPages{Science education}{84}{3}{287--312,}
\newblock

\newblock

\PrintBackRefs{\CurrentBib}

\bibitem [\protect \citeauthoryear {%
Fiacco%
, Jiang%
, Adamson%
\BCBL {}\ \BBA {} Ros{\'e}%
}{%
Fiacco%
\ \protect \BOthers {.}}{%
{\protect \APACyear {2022}}%
}]{%
fiacco-etal-2022-toward}
\APACinsertmetastar {%
fiacco-etal-2022-toward}%
\begin{APACrefauthors}%
Fiacco, J.%
, Jiang, S.%
, Adamson, D.%
\BCBL {} Ros{\'e}, C.%
\end{APACrefauthors}%
\unskip\
\newblock
\APACrefYearMonthDay{2022}{{\APACmonth{07}}}{}.
\newblock
{\BBOQ}\APACrefatitle {Toward Automatic Discourse Parsing of Student Writing Motivated by Neural Interpretation} {Toward automatic discourse parsing of student writing motivated by neural interpretation}.{\BBCQ}
\newblock
 E.~Kochmar\ \BOthers {.}\ (\BEDS), \APACrefbtitle {Proceedings of the 17th Workshop on Innovative Use of NLP for Building Educational Applications (BEA 2022)} {Proceedings of the 17th workshop on innovative use of nlp for building educational applications (bea 2022)}\ (\BPGS\ 204--215).
\newblock
\APACaddressPublisher{Seattle, Washington}{Association for Computational Linguistics}.
\newblock
\begin{APACrefURL} {https://aclanthology.org/2022.bea-1.25} \end{APACrefURL}
\PrintBackRefs{\CurrentBib}

\bibitem [\protect \citeauthoryear {%
Gombert%
\ \protect \BOthers {.}}{%
Gombert%
\ \protect \BOthers {.}}{%
{\protect \APACyear {2023}}%
}]{%
gombert2023coding}
\APACinsertmetastar {%
gombert2023coding}%
\begin{APACrefauthors}%
Gombert, S.%
, Di~Mitri, D.%
, Karademir, O.%
, Kubsch, M.%
, Kolbe, H.%
, Tautz, S.%
\BDBL {}Drachsler, H.%
\end{APACrefauthors}%
\unskip\
\newblock
\APACrefYearMonthDay{2023}{}{}.
\newblock
{\BBOQ}\APACrefatitle {Coding energy knowledge in constructed responses with explainable NLP models} {Coding energy knowledge in constructed responses with explainable nlp models}.{\BBCQ}
\newblock
\APACjournalVolNumPages{Journal of Computer Assisted Learning}{39}{3}{767--786,}
\newblock

\newblock

\PrintBackRefs{\CurrentBib}

\bibitem [\protect \citeauthoryear {%
Ha%
, H~Nehm%
, Urban-Lurain%
\BCBL {}\ \BBA {} E.%
}{%
Ha%
\ \protect \BOthers {.}}{%
{\protect \APACyear {2011}}%
}]{%
Ha2011computerized}
\APACinsertmetastar {%
Ha2011computerized}%
\begin{APACrefauthors}%
Ha, M.%
, H~Nehm, R.%
, Urban-Lurain, M.%
\BCBL {} E., M.J.%
\end{APACrefauthors}%
\unskip\
\newblock
\APACrefYearMonthDay{2011}{}{}.
\newblock
{\BBOQ}\APACrefatitle {Applying computerized-scoring models of written biological explanations across courses and colleges: prospects and limitations} {Applying computerized-scoring models of written biological explanations across courses and colleges: prospects and limitations}.{\BBCQ}
\newblock
\APACjournalVolNumPages{CBE—Life Sciences Education}{10}{4}{379-393,}
\newblock

\newblock

\PrintBackRefs{\CurrentBib}

\bibitem [\protect \citeauthoryear {%
Haudek%
\ \BBA {} Zhai%
}{%
Haudek%
\ \BBA {} Zhai%
}{%
{\protect \APACyear {2023}}%
}]{%
haudek2023examining}
\APACinsertmetastar {%
haudek2023examining}%
\begin{APACrefauthors}%
Haudek, K.C.%
\BCBT {}\ \BBA {} Zhai, X.%
\end{APACrefauthors}%
\unskip\
\newblock
\APACrefYearMonthDay{2023}{}{}.
\newblock
{\BBOQ}\APACrefatitle {Examining the Effect of Assessment Construct Characteristics on Machine Learning Scoring of Scientific Argumentation} {Examining the effect of assessment construct characteristics on machine learning scoring of scientific argumentation}.{\BBCQ}
\newblock
\APACjournalVolNumPages{International Journal of Artificial Intelligence in Education}{}{}{1--28,}
\newblock

\newblock

\PrintBackRefs{\CurrentBib}

\bibitem [\protect \citeauthoryear {%
He%
\ \protect \BOthers {.}}{%
He%
\ \protect \BOthers {.}}{%
{\protect \APACyear {2023}}%
}]{%
zhang2023assessing}
\APACinsertmetastar {%
zhang2023assessing}%
\begin{APACrefauthors}%
He, Z.%
, Chuhao, W.%
, Jingyi, X.%
, Lyu, Y.%
, Jie, C.%
\BCBL {} Carroll, J.M.%
\end{APACrefauthors}%
\unskip\
\newblock
\APACrefYearMonthDay{2023}{}{}.
\newblock
{\BBOQ}\APACrefatitle {Redefining Qualitative Analysis in the AI Era: Utilizing ChatGPT for Efficient Thematic Analysis} {Redefining qualitative analysis in the ai era: Utilizing chatgpt for efficient thematic analysis}.{\BBCQ}
\newblock
\APACjournalVolNumPages{arXiv preprint arXiv:2309.10771}{}{}{,}
\newblock

\newblock

\PrintBackRefs{\CurrentBib}

\bibitem [\protect \citeauthoryear {%
Krell%
, Vorholzer%
\BCBL {}\ \BBA {} Nehring%
}{%
Krell%
\ \protect \BOthers {.}}{%
{\protect \APACyear {2022}}%
}]{%
krell2022scientific}
\APACinsertmetastar {%
krell2022scientific}%
\begin{APACrefauthors}%
Krell, M.%
, Vorholzer, A.%
\BCBL {} Nehring, A.%
\end{APACrefauthors}%
\unskip\
\newblock
\APACrefYearMonthDay{2022}{}{}.
\newblock
{\BBOQ}\APACrefatitle {Scientific Reasoning in Science Education: From Global Measures to Fine-Grained Descriptions of Students’ Competencies. Educ. Sci. 2022, 12, 97} {Scientific reasoning in science education: From global measures to fine-grained descriptions of students’ competencies. educ. sci. 2022, 12, 97}.{\BBCQ}
\newblock
\APACjournalVolNumPages{Scientific Reasoning in Science Education}{}{}{,}
\newblock

\newblock

\PrintBackRefs{\CurrentBib}

\bibitem [\protect \citeauthoryear {%
Kubsch%
\ \protect \BOthers {.}}{%
Kubsch%
\ \protect \BOthers {.}}{%
{\protect \APACyear {2022}}%
}]{%
kubsch2022toward}
\APACinsertmetastar {%
kubsch2022toward}%
\begin{APACrefauthors}%
Kubsch, M.%
, Czinczel, B.%
, Lossjew, J.%
, Wyrwich, T.%
, Bednorz, D.%
, Bernholt, S.%
\BDBL {}others%
\end{APACrefauthors}%
\unskip\
\newblock
\APACrefYearMonthDay{2022}{}{}.
\newblock
{\BBOQ}\APACrefatitle {Toward learning progression analytics—Developing learning environments for the automated analysis of learning using evidence centered design} {Toward learning progression analytics—developing learning environments for the automated analysis of learning using evidence centered design}.{\BBCQ}
\newblock
 \APACrefbtitle {Frontiers in education} {Frontiers in education}\ (\BVOL~7, \BPG~981910).
\PrintBackRefs{\CurrentBib}

\bibitem [\protect \citeauthoryear {%
Kumar%
\ \BBA {} Boulanger%
}{%
Kumar%
\ \BBA {} Boulanger%
}{%
{\protect \APACyear {2021}}%
}]{%
kumar2021automated}
\APACinsertmetastar {%
kumar2021automated}%
\begin{APACrefauthors}%
Kumar, V.S.%
\BCBT {}\ \BBA {} Boulanger, D.%
\end{APACrefauthors}%
\unskip\
\newblock
\APACrefYearMonthDay{2021}{}{}.
\newblock
{\BBOQ}\APACrefatitle {Automated essay scoring and the deep learning black box: How are rubric scores determined?} {Automated essay scoring and the deep learning black box: How are rubric scores determined?}{\BBCQ}
\newblock
\APACjournalVolNumPages{International Journal of Artificial Intelligence in Education}{31}{}{538--584,}
\newblock

\newblock

\PrintBackRefs{\CurrentBib}

\bibitem [\protect \citeauthoryear {%
Kwon%
\ \BBA {} Lawson%
}{%
Kwon%
\ \BBA {} Lawson%
}{%
{\protect \APACyear {2000}}%
}]{%
kwon2000linking}
\APACinsertmetastar {%
kwon2000linking}%
\begin{APACrefauthors}%
Kwon, Y\BHBI J.%
\BCBT {}\ \BBA {} Lawson, A.E.%
\end{APACrefauthors}%
\unskip\
\newblock
\APACrefYearMonthDay{2000}{}{}.
\newblock
{\BBOQ}\APACrefatitle {Linking brain growth with the development of scientific reasoning ability and conceptual change during adolescence} {Linking brain growth with the development of scientific reasoning ability and conceptual change during adolescence}.{\BBCQ}
\newblock
\APACjournalVolNumPages{Journal of Research in Science Teaching: The Official Journal of the National Association for Research in Science Teaching}{37}{1}{44--62,}
\newblock

\newblock

\PrintBackRefs{\CurrentBib}

\bibitem [\protect \citeauthoryear {%
Lamb%
, Hand%
\BCBL {}\ \BBA {} Kavner%
}{%
Lamb%
\ \protect \BOthers {.}}{%
{\protect \APACyear {2021}}%
}]{%
lamb2021computational}
\APACinsertmetastar {%
lamb2021computational}%
\begin{APACrefauthors}%
Lamb, R.%
, Hand, B.%
\BCBL {} Kavner, A.%
\end{APACrefauthors}%
\unskip\
\newblock
\APACrefYearMonthDay{2021}{}{}.
\newblock
{\BBOQ}\APACrefatitle {Computational modeling of the effects of the science writing heuristic on student critical thinking in science using machine learning} {Computational modeling of the effects of the science writing heuristic on student critical thinking in science using machine learning}.{\BBCQ}
\newblock
\APACjournalVolNumPages{Journal of Science Education and Technology}{30}{}{283--297,}
\newblock

\newblock

\PrintBackRefs{\CurrentBib}

\bibitem [\protect \citeauthoryear {%
Latif%
\ \BBA {} Zhai%
}{%
Latif%
\ \BBA {} Zhai%
}{%
{\protect \APACyear {2024}}%
}]{%
latif2024fine}
\APACinsertmetastar {%
latif2024fine}%
\begin{APACrefauthors}%
Latif, E.%
\BCBT {}\ \BBA {} Zhai, X.%
\end{APACrefauthors}%
\unskip\
\newblock
\APACrefYearMonthDay{2024}{}{}.
\newblock
{\BBOQ}\APACrefatitle {Fine-tuning chatgpt for automatic scoring} {Fine-tuning chatgpt for automatic scoring}.{\BBCQ}
\newblock
\APACjournalVolNumPages{Computers and Education: Artificial Intelligence}{}{}{100210,}
\newblock

\newblock

\PrintBackRefs{\CurrentBib}

\bibitem [\protect \citeauthoryear {%
H.~Li%
, Gobert%
\BCBL {}\ \BBA {} Dickler%
}{%
H.~Li%
\ \protect \BOthers {.}}{%
{\protect \APACyear {2017}}%
}]{%
Liqutomated}
\APACinsertmetastar {%
Liqutomated}%
\begin{APACrefauthors}%
Li, H.%
, Gobert, J.%
\BCBL {} Dickler, R.%
\end{APACrefauthors}%
\unskip\
\newblock
\APACrefYearMonthDay{2017}{01}{}.
\newblock
{\BBOQ}\APACrefatitle {Automated Assessment for Scientific Explanations in On-line Science Inquiry} {Automated assessment for scientific explanations in on-line science inquiry}.{\BBCQ}.
\PrintBackRefs{\CurrentBib}

\bibitem [\protect \citeauthoryear {%
X.~Li%
\ \protect \BOthers {.}}{%
X.~Li%
\ \protect \BOthers {.}}{%
{\protect \APACyear {2019}}%
}]{%
li2019word}
\APACinsertmetastar {%
li2019word}%
\begin{APACrefauthors}%
Li, X.%
, Meng, Y.%
, Sun, X.%
, Han, Q.%
, Yuan, A.%
\BCBL {} Li, J.%
\end{APACrefauthors}%
\unskip\
\newblock
\APACrefYearMonthDay{2019}{}{}.
\newblock
{\BBOQ}\APACrefatitle {Is word segmentation necessary for deep learning of Chinese representations?} {Is word segmentation necessary for deep learning of chinese representations?}{\BBCQ}
\newblock
\APACjournalVolNumPages{arXiv preprint arXiv:1905.05526}{}{}{,}
\newblock

\newblock

\PrintBackRefs{\CurrentBib}

\bibitem [\protect \citeauthoryear {%
Liao%
, Chen-Huei%
, Kuo%
\BCBL {}\ \BBA {} Pai%
}{%
Liao%
\ \protect \BOthers {.}}{%
{\protect \APACyear {2012}}%
}]{%
liao2012effectivenesss}
\APACinsertmetastar {%
liao2012effectivenesss}%
\begin{APACrefauthors}%
Liao%
, Chen-Huei%
, Kuo, B\BHBI C.%
\BCBL {} Pai, K\BHBI C.%
\end{APACrefauthors}%
\unskip\
\newblock
\APACrefYearMonthDay{2012}{}{}.
\newblock
{\BBOQ}\APACrefatitle {Effectiveness of Automated Chinese Sentence Scoring with Latent Semantic Analysis} {Effectiveness of automated chinese sentence scoring with latent semantic analysis}.{\BBCQ}
\newblock
\APACjournalVolNumPages{Turkish Online Journal of Educational Technology-TOJET}{11}{2}{80-87,}
\newblock

\newblock

\PrintBackRefs{\CurrentBib}

\bibitem [\protect \citeauthoryear {%
Limna%
, Jakwatanatham%
, Siripipattanakul%
, Kaewpuang%
\BCBL {}\ \BBA {} Sriboonruang%
}{%
Limna%
\ \protect \BOthers {.}}{%
{\protect \APACyear {2022}}%
}]{%
limna2022review}
\APACinsertmetastar {%
limna2022review}%
\begin{APACrefauthors}%
Limna, P.%
, Jakwatanatham, S.%
, Siripipattanakul, S.%
, Kaewpuang, P.%
\BCBL {} Sriboonruang, P.%
\end{APACrefauthors}%
\unskip\
\newblock
\APACrefYearMonthDay{2022}{}{}.
\newblock
{\BBOQ}\APACrefatitle {A review of artificial intelligence (AI) in education during the digital era} {A review of artificial intelligence (ai) in education during the digital era}.{\BBCQ}
\newblock
\APACjournalVolNumPages{Advance Knowledge for Executives}{1}{1}{1--9,}
\newblock

\newblock

\PrintBackRefs{\CurrentBib}

\bibitem [\protect \citeauthoryear {%
Liu%
, He%
, Liu%
, Liu%
\BCBL {}\ \BBA {} Zhai%
}{%
Liu%
\ \protect \BOthers {.}}{%
{\protect \APACyear {2023}}%
}]{%
liu2023context}
\APACinsertmetastar {%
liu2023context}%
\begin{APACrefauthors}%
Liu, Z.%
, He, X.%
, Liu, L.%
, Liu, T.%
\BCBL {} Zhai, X.%
\end{APACrefauthors}%
\unskip\
\newblock
\APACrefYearMonthDay{2023}{}{}.
\newblock
{\BBOQ}\APACrefatitle {Context matters: A strategy to pre-train language model for science education} {Context matters: A strategy to pre-train language model for science education}.{\BBCQ}
\newblock
\APACjournalVolNumPages{arXiv preprint arXiv:2301.12031}{}{}{,}
\newblock

\newblock

\PrintBackRefs{\CurrentBib}

\bibitem [\protect \citeauthoryear {%
Lockwood%
}{%
Lockwood%
}{%
{\protect \APACyear {2015}}%
}]{%
lockwood2015handbook}
\APACinsertmetastar {%
lockwood2015handbook}%
\begin{APACrefauthors}%
Lockwood, J.%
\end{APACrefauthors}%
\unskip\
\newblock
\APACrefYearMonthDay{2015}{}{}.
\newblock
{\BBOQ}\APACrefatitle {Handbook of automated essay evaluation: Current applications and new directions} {Handbook of automated essay evaluation: Current applications and new directions}.{\BBCQ}
\newblock
\APACjournalVolNumPages{Writing \& Pedagogy}{6}{}{437--441,}
\newblock

\newblock

\PrintBackRefs{\CurrentBib}

\bibitem [\protect \citeauthoryear {%
McNeill%
\ \BBA {} Krajcik%
}{%
McNeill%
\ \BBA {} Krajcik%
}{%
{\protect \APACyear {2008}}%
}]{%
mcneill2008inquiry}
\APACinsertmetastar {%
mcneill2008inquiry}%
\begin{APACrefauthors}%
McNeill, K.L.%
\BCBT {}\ \BBA {} Krajcik, J.%
\end{APACrefauthors}%
\unskip\
\newblock
\APACrefYearMonthDay{2008}{}{}.
\newblock
{\BBOQ}\APACrefatitle {Inquiry and scientific explanations: Helping students use evidence and reasoning} {Inquiry and scientific explanations: Helping students use evidence and reasoning}.{\BBCQ}
\newblock
\APACjournalVolNumPages{Science as inquiry in the secondary setting}{121}{}{34,}
\newblock

\newblock

\PrintBackRefs{\CurrentBib}

\bibitem [\protect \citeauthoryear {%
Mislevy%
, Yan%
, Gobert%
\BCBL {}\ \BBA {} Sao~Pedro%
}{%
Mislevy%
\ \protect \BOthers {.}}{%
{\protect \APACyear {2020}}%
}]{%
mislevy2020automated}
\APACinsertmetastar {%
mislevy2020automated}%
\begin{APACrefauthors}%
Mislevy, R.J.%
, Yan, D.%
, Gobert, J.%
\BCBL {} Sao~Pedro, M.%
\end{APACrefauthors}%
\unskip\
\newblock
\APACrefYearMonthDay{2020}{}{}.
\newblock
{\BBOQ}\APACrefatitle {Automated scoring in intelligent tutoring systems} {Automated scoring in intelligent tutoring systems}.{\BBCQ}
\newblock
 \APACrefbtitle {Handbook of automated scoring} {Handbook of automated scoring}\ (\BPGS\ 403--422).
\newblock
\APACaddressPublisher{}{Chapman and Hall/CRC}.
\PrintBackRefs{\CurrentBib}

\bibitem [\protect \citeauthoryear {%
Moharreri%
, Ha%
\BCBL {}\ \BBA {} Nehm%
}{%
Moharreri%
\ \protect \BOthers {.}}{%
{\protect \APACyear {2014}}%
}]{%
Moharreri2014EvoGrader}
\APACinsertmetastar {%
Moharreri2014EvoGrader}%
\begin{APACrefauthors}%
Moharreri, K.%
, Ha, M.%
\BCBL {} Nehm, R.H.%
\end{APACrefauthors}%
\unskip\
\newblock
\APACrefYearMonthDay{2014}{}{}.
\newblock
{\BBOQ}\APACrefatitle {EvoGrader: an online formative assessment tool for automatically evaluating written evolutionary explanations} {Evograder: an online formative assessment tool for automatically evaluating written evolutionary explanations}.{\BBCQ}
\newblock
\APACjournalVolNumPages{Evolution: Education and Outreach}{7}{}{1--14,}
\newblock

\newblock

\PrintBackRefs{\CurrentBib}

\bibitem [\protect \citeauthoryear {%
Nehm%
, Ha%
\BCBL {}\ \BBA {} Mayfield%
}{%
Nehm%
\ \protect \BOthers {.}}{%
{\protect \APACyear {2012}}%
}]{%
Nehm2012Transforming}
\APACinsertmetastar {%
Nehm2012Transforming}%
\begin{APACrefauthors}%
Nehm, R.%
, Ha, M.%
\BCBL {} Mayfield, E.%
\end{APACrefauthors}%
\unskip\
\newblock
\APACrefYearMonthDay{2012}{}{}.
\newblock
{\BBOQ}\APACrefatitle {Transforming biology assessment with machine learning: automated scoring of written evolutionary explanations} {Transforming biology assessment with machine learning: automated scoring of written evolutionary explanations}.{\BBCQ}
\newblock
\APACjournalVolNumPages{Journal of Science Education and Technology}{21}{}{183-196.,}
\newblock

\newblock

\PrintBackRefs{\CurrentBib}

\bibitem [\protect \citeauthoryear {%
Neri%
\ \BBA {} Retelsdorf%
}{%
Neri%
\ \BBA {} Retelsdorf%
}{%
{\protect \APACyear {2022}}%
}]{%
neri2022role}
\APACinsertmetastar {%
neri2022role}%
\begin{APACrefauthors}%
Neri, N.C.%
\BCBT {}\ \BBA {} Retelsdorf, J.%
\end{APACrefauthors}%
\unskip\
\newblock
\APACrefYearMonthDay{2022}{}{}.
\newblock
{\BBOQ}\APACrefatitle {The role of linguistic features in science and math comprehension and performance: A systematic review and desiderata for future research} {The role of linguistic features in science and math comprehension and performance: A systematic review and desiderata for future research}.{\BBCQ}
\newblock
\APACjournalVolNumPages{Educational Research Review}{36}{}{100460,}
\newblock

\newblock

\PrintBackRefs{\CurrentBib}

\bibitem [\protect \citeauthoryear {%
OECD%
}{%
OECD%
}{%
{\protect \APACyear {2019}}%
}]{%
OECD2019}
\APACinsertmetastar {%
OECD2019}%
\begin{APACrefauthors}%
OECD%
\end{APACrefauthors}%
\unskip\
\newblock
\APACrefYearMonthDay{2019}{}{}.
\newblock
\APACrefbtitle {PISA2018 Well-being Framework} {Pisa2018 well-being framework}\ \APACbVolEdTR{}{\BTR{}}.
\newblock
\APACaddressInstitutionEqAuth{}{OECD}.
\PrintBackRefs{\CurrentBib}

\bibitem [\protect \citeauthoryear {%
Ormerod%
\ \protect \BOthers {.}}{%
Ormerod%
\ \protect \BOthers {.}}{%
{\protect \APACyear {2023}}%
}]{%
ormerod2023automated}
\APACinsertmetastar {%
ormerod2023automated}%
\begin{APACrefauthors}%
Ormerod, C.%
, Lottridge, S.%
, Harris, A.E.%
, Patel, M.%
, van Wamelen, P.%
, Kodeswaran, B.%
\BDBL {}Young, M.%
\end{APACrefauthors}%
\unskip\
\newblock
\APACrefYearMonthDay{2023}{}{}.
\newblock
{\BBOQ}\APACrefatitle {Automated short answer scoring using an ensemble of neural networks and latent semantic analysis classifiers} {Automated short answer scoring using an ensemble of neural networks and latent semantic analysis classifiers}.{\BBCQ}
\newblock
\APACjournalVolNumPages{International Journal of Artificial Intelligence in Education}{33}{3}{467--496,}
\newblock

\newblock

\PrintBackRefs{\CurrentBib}

\bibitem [\protect \citeauthoryear {%
Reiser%
, Berland%
\BCBL {}\ \BBA {} Kenyon%
}{%
Reiser%
\ \protect \BOthers {.}}{%
{\protect \APACyear {2012}}%
}]{%
reiser2012engaging}
\APACinsertmetastar {%
reiser2012engaging}%
\begin{APACrefauthors}%
Reiser, B.J.%
, Berland, L.K.%
\BCBL {} Kenyon, L.%
\end{APACrefauthors}%
\unskip\
\newblock
\APACrefYearMonthDay{2012}{}{}.
\newblock
{\BBOQ}\APACrefatitle {Engaging students in the scientific practices of explanation and argumentation} {Engaging students in the scientific practices of explanation and argumentation}.{\BBCQ}
\newblock
\APACjournalVolNumPages{The Science Teacher}{79}{4}{34,}
\newblock

\newblock

\PrintBackRefs{\CurrentBib}

\bibitem [\protect \citeauthoryear {%
Riordan%
\ \protect \BOthers {.}}{%
Riordan%
\ \protect \BOthers {.}}{%
{\protect \APACyear {2020}}%
}]{%
riordan2020empirical}
\APACinsertmetastar {%
riordan2020empirical}%
\begin{APACrefauthors}%
Riordan, B.%
, Bichler, S.%
, Bradford, A.%
, Chen, J.K.%
, Wiley, K.%
, Gerard, L.%
\BCBL {} Linn, M.C.%
\end{APACrefauthors}%
\unskip\
\newblock
\APACrefYearMonthDay{2020}{}{}.
\newblock
{\BBOQ}\APACrefatitle {An empirical investigation of neural methods for content scoring of science explanations} {An empirical investigation of neural methods for content scoring of science explanations}.{\BBCQ}
\newblock
 \APACrefbtitle {Proceedings of the fifteenth workshop on innovative use of NLP for building educational applications} {Proceedings of the fifteenth workshop on innovative use of nlp for building educational applications}\ (\BPGS\ 135--144).
\PrintBackRefs{\CurrentBib}

\bibitem [\protect \citeauthoryear {%
Schober%
, Boer%
\BCBL {}\ \BBA {} Schwarte%
}{%
Schober%
\ \protect \BOthers {.}}{%
{\protect \APACyear {2018}}%
}]{%
schober2018correlation}
\APACinsertmetastar {%
schober2018correlation}%
\begin{APACrefauthors}%
Schober, P.%
, Boer, C.%
\BCBL {} Schwarte, L.A.%
\end{APACrefauthors}%
\unskip\
\newblock
\APACrefYearMonthDay{2018}{}{}.
\newblock
{\BBOQ}\APACrefatitle {Correlation coefficients: appropriate use and interpretation} {Correlation coefficients: appropriate use and interpretation}.{\BBCQ}
\newblock
\APACjournalVolNumPages{Anesthesia \& analgesia}{126}{5}{1763--1768,}
\newblock

\newblock

\PrintBackRefs{\CurrentBib}

\bibitem [\protect \citeauthoryear {%
SUN%
\ \BBA {} TIAN%
}{%
SUN%
\ \BBA {} TIAN%
}{%
{\protect \APACyear {2017}}%
}]{%
sun2017cultural}
\APACinsertmetastar {%
sun2017cultural}%
\begin{APACrefauthors}%
SUN, M.%
\BCBT {}\ \BBA {} TIAN, Z\BHBI x.%
\end{APACrefauthors}%
\unskip\
\newblock
\APACrefYearMonthDay{2017}{}{}.
\newblock
{\BBOQ}\APACrefatitle {The cultural differences between English and Chinese courtesy languages} {The cultural differences between english and chinese courtesy languages}.{\BBCQ}
\newblock
\APACjournalVolNumPages{Journal of Literature and Art Studies}{7}{3}{340--344,}
\newblock

\newblock

\PrintBackRefs{\CurrentBib}

\bibitem [\protect \citeauthoryear {%
Vosniadou%
}{%
Vosniadou%
}{%
{\protect \APACyear {2012}}%
}]{%
Vosniadou2019understanding}
\APACinsertmetastar {%
Vosniadou2019understanding}%
\begin{APACrefauthors}%
Vosniadou, S.%
\end{APACrefauthors}%
\unskip\
\newblock
\APACrefYearMonthDay{2012}{}{}.
\newblock
{\BBOQ}\APACrefatitle {The development of students' understanding of science} {The development of students' understanding of science}.{\BBCQ}
\newblock
\APACjournalVolNumPages{In Frontiers in Education}{4}{}{32,}
\newblock

\newblock

\PrintBackRefs{\CurrentBib}

\bibitem [\protect \citeauthoryear {%
C.~Wang%
, Liu%
, Wang%
, Sun%
\BCBL {}\ \BBA {} zhang%
}{%
C.~Wang%
\ \protect \BOthers {.}}{%
{\protect \APACyear {2021}}%
}]{%
Wang2021automated}
\APACinsertmetastar {%
Wang2021automated}%
\begin{APACrefauthors}%
Wang, C.%
, Liu, X.%
, Wang, L.%
, Sun, Y.%
\BCBL {} zhang, H.%
\end{APACrefauthors}%
\unskip\
\newblock
\APACrefYearMonthDay{2021}{}{}.
\newblock
{\BBOQ}\APACrefatitle {Automated scoring of Chinese grades 7–9 students’ competence in interpreting and arguing from evidence} {Automated scoring of chinese grades 7–9 students’ competence in interpreting and arguing from evidence}.{\BBCQ}
\newblock
\APACjournalVolNumPages{Journal of Science Education and Technology}{30}{}{269-282,}
\newblock

\newblock

\PrintBackRefs{\CurrentBib}

\bibitem [\protect \citeauthoryear {%
Y.~Wang%
\ \BBA {} Chen%
}{%
Y.~Wang%
\ \BBA {} Chen%
}{%
{\protect \APACyear {2013}}%
}]{%
wang2013differences}
\APACinsertmetastar {%
wang2013differences}%
\begin{APACrefauthors}%
Wang, Y.%
\BCBT {}\ \BBA {} Chen, J.%
\end{APACrefauthors}%
\unskip\
\newblock
\APACrefYearMonthDay{2013}{}{}.
\newblock
{\BBOQ}\APACrefatitle {Differences of English and Chinese as Written Languages and Strategies in English Writing Teaching.} {Differences of english and chinese as written languages and strategies in english writing teaching.}{\BBCQ}
\newblock
\APACjournalVolNumPages{Theory \& Practice in Language Studies (TPLS)}{3}{4}{,}
\newblock

\newblock

\PrintBackRefs{\CurrentBib}

\bibitem [\protect \citeauthoryear {%
Williams%
\ \BBA {} Bever%
}{%
Williams%
\ \BBA {} Bever%
}{%
{\protect \APACyear {2010}}%
}]{%
Williamsf2010Chinese}
\APACinsertmetastar {%
Williamsf2010Chinese}%
\begin{APACrefauthors}%
Williams, C.%
\BCBT {}\ \BBA {} Bever, T.%
\end{APACrefauthors}%
\unskip\
\newblock
\APACrefYearMonthDay{2010}{}{}.
\newblock
{\BBOQ}\APACrefatitle {Chinese character decoding: a semantic bias?} {Chinese character decoding: a semantic bias?}{\BBCQ}
\newblock
\APACjournalVolNumPages{Reading and Writing}{}{}{589-605,}
\newblock

\newblock

\PrintBackRefs{\CurrentBib}

\bibitem [\protect \citeauthoryear {%
Wu%
\ \protect \BOthers {.}}{%
Wu%
\ \protect \BOthers {.}}{%
{\protect \APACyear {2023}}%
}]{%
Wu2023GPTOverview}
\APACinsertmetastar {%
Wu2023GPTOverview}%
\begin{APACrefauthors}%
Wu, T.%
, He, S.%
, Liu, J.%
, Sun, S.%
, Liu, K.%
, Han, Q\BHBI L.%
\BCBL {} Tang, Y.%
\end{APACrefauthors}%
\unskip\
\newblock
\APACrefYearMonthDay{2023}{}{}.
\newblock
{\BBOQ}\APACrefatitle {A Brief Overview of ChatGPT: The History, Status Quo and Potential Future Development} {A brief overview of chatgpt: The history, status quo and potential future development}.{\BBCQ}
\newblock
\APACjournalVolNumPages{IEEE/CAA Journal of Automatica Sinica}{10}{}{1122-1136,}
\newblock

\newblock

\PrintBackRefs{\CurrentBib}

\bibitem [\protect \citeauthoryear {%
Yang%
, Zhang%
\BCBL {}\ \BBA {} Liang%
}{%
Yang%
\ \protect \BOthers {.}}{%
{\protect \APACyear {2018}}%
}]{%
yang2018subword}
\APACinsertmetastar {%
yang2018subword}%
\begin{APACrefauthors}%
Yang, J.%
, Zhang, Y.%
\BCBL {} Liang, S.%
\end{APACrefauthors}%
\unskip\
\newblock
\APACrefYearMonthDay{2018}{}{}.
\newblock
{\BBOQ}\APACrefatitle {Subword encoding in lattice LSTM for Chinese word segmentation} {Subword encoding in lattice lstm for chinese word segmentation}.{\BBCQ}
\newblock
\APACjournalVolNumPages{arXiv preprint arXiv:1810.12594}{}{}{,}
\newblock

\newblock

\PrintBackRefs{\CurrentBib}

\bibitem [\protect \citeauthoryear {%
Yao%
\ \BBA {} Guo%
}{%
Yao%
\ \BBA {} Guo%
}{%
{\protect \APACyear {2018}}%
}]{%
Yao2018explanation}
\APACinsertmetastar {%
Yao2018explanation}%
\begin{APACrefauthors}%
Yao, J.%
\BCBT {}\ \BBA {} Guo, Y.%
\end{APACrefauthors}%
\unskip\
\newblock
\APACrefYearMonthDay{2018}{}{}.
\newblock
{\BBOQ}\APACrefatitle {Validity evidence for a learning progression of scientific explanation} {Validity evidence for a learning progression of scientific explanation}.{\BBCQ}
\newblock
\APACjournalVolNumPages{Journal of Research in Science Teaching}{55}{2}{299-317,}
\newblock

\newblock

\PrintBackRefs{\CurrentBib}

\bibitem [\protect \citeauthoryear {%
Zhai%
}{%
Zhai%
}{%
{\protect \APACyear {2022}}%
}]{%
zhai2022assessing}
\APACinsertmetastar {%
zhai2022assessing}%
\begin{APACrefauthors}%
Zhai, X.%
\end{APACrefauthors}%
\unskip\
\newblock
\APACrefYearMonthDay{2022}{}{}.
\newblock
{\BBOQ}\APACrefatitle {Assessing high-school students' modeling performance on Newtonian mechanics} {Assessing high-school students' modeling performance on newtonian mechanics}.{\BBCQ}
\newblock
\APACjournalVolNumPages{Journal of Research in Science Teaching}{59}{8}{1313--1353,}
\newblock

\newblock

\PrintBackRefs{\CurrentBib}

\bibitem [\protect \citeauthoryear {%
Zhai%
}{%
Zhai%
}{%
{\protect \APACyear {2023}}%
}]{%
zhai2023chatgpt}
\APACinsertmetastar {%
zhai2023chatgpt}%
\begin{APACrefauthors}%
Zhai, X.%
\end{APACrefauthors}%
\unskip\
\newblock
\APACrefYearMonthDay{2023}{}{}.
\newblock
{\BBOQ}\APACrefatitle {Chatgpt for next generation science learning} {Chatgpt for next generation science learning}.{\BBCQ}
\newblock
\APACjournalVolNumPages{XRDS: Crossroads, The ACM Magazine for Students}{29}{3}{42--46,}
\newblock

\newblock

\PrintBackRefs{\CurrentBib}

\bibitem [\protect \citeauthoryear {%
Zhai%
}{%
Zhai%
}{%
{\protect \APACyear {2024}}%
}]{%
zhai2024ai}
\APACinsertmetastar {%
zhai2024ai}%
\begin{APACrefauthors}%
Zhai, X.%
\end{APACrefauthors}%
\unskip\
\newblock
\APACrefYearMonthDay{2024}{}{}.
\newblock
{\BBOQ}\APACrefatitle {AI and Machine Learning for Next Generation Sci-ence Assessments} {Ai and machine learning for next generation sci-ence assessments}.{\BBCQ}
\newblock
\APACjournalVolNumPages{Machine Learning, Natural Language Processing, and Psychometrics}{}{}{201,}
\newblock

\newblock

\PrintBackRefs{\CurrentBib}

\bibitem [\protect \citeauthoryear {%
Zhai%
\ \BBA {} Nehm%
}{%
Zhai%
\ \BBA {} Nehm%
}{%
{\protect \APACyear {2023}}%
}]{%
zhai2023ai}
\APACinsertmetastar {%
zhai2023ai}%
\begin{APACrefauthors}%
Zhai, X.%
\BCBT {}\ \BBA {} Nehm, R.H.%
\end{APACrefauthors}%
\unskip\
\newblock
\APACrefYearMonthDay{2023}{}{}.
\newblock
{\BBOQ}\APACrefatitle {AI and formative assessment: The train has left the station} {Ai and formative assessment: The train has left the station}.{\BBCQ}
\newblock
\APACjournalVolNumPages{Journal of Research in Science Teaching}{}{}{,}
\newblock

\newblock

\PrintBackRefs{\CurrentBib}

\bibitem [\protect \citeauthoryear {%
Zhai%
\ \BBA {} Pellegrino%
}{%
Zhai%
\ \BBA {} Pellegrino%
}{%
{\protect \APACyear {2023}}%
}]{%
zhai2023large}
\APACinsertmetastar {%
zhai2023large}%
\begin{APACrefauthors}%
Zhai, X.%
\BCBT {}\ \BBA {} Pellegrino, J.W.%
\end{APACrefauthors}%
\unskip\
\newblock
\APACrefYearMonthDay{2023}{}{}.
\newblock
{\BBOQ}\APACrefatitle {Large-scale assessment in science education} {Large-scale assessment in science education}.{\BBCQ}
\newblock
 \APACrefbtitle {Handbook of research on science education} {Handbook of research on science education}\ (\BPGS\ 1045--1097).
\newblock
\APACaddressPublisher{}{Routledge}.
\PrintBackRefs{\CurrentBib}

\bibitem [\protect \citeauthoryear {%
Zhai%
, Shi%
\BCBL {}\ \BBA {} Nehm%
}{%
Zhai%
\ \protect \BOthers {.}}{%
{\protect \APACyear {2021}}%
}]{%
zhai2021meta}
\APACinsertmetastar {%
zhai2021meta}%
\begin{APACrefauthors}%
Zhai, X.%
, Shi, L.%
\BCBL {} Nehm, R.H.%
\end{APACrefauthors}%
\unskip\
\newblock
\APACrefYearMonthDay{2021}{}{}.
\newblock
{\BBOQ}\APACrefatitle {A meta-analysis of machine learning-based science assessments: Factors impacting machine-human score agreements} {A meta-analysis of machine learning-based science assessments: Factors impacting machine-human score agreements}.{\BBCQ}
\newblock
\APACjournalVolNumPages{Journal of Science Education and Technology}{30}{}{361--379,}
\newblock

\newblock

\PrintBackRefs{\CurrentBib}

\bibitem [\protect \citeauthoryear {%
Zhai%
, Yin%
, Pellegrino%
, Haudek%
\BCBL {}\ \BBA {} Shi%
}{%
Zhai%
\ \protect \BOthers {.}}{%
{\protect \APACyear {2020}}%
}]{%
zhai2020applying}
\APACinsertmetastar {%
zhai2020applying}%
\begin{APACrefauthors}%
Zhai, X.%
, Yin, Y.%
, Pellegrino, J.W.%
, Haudek, K.C.%
\BCBL {} Shi, L.%
\end{APACrefauthors}%
\unskip\
\newblock
\APACrefYearMonthDay{2020}{}{}.
\newblock
{\BBOQ}\APACrefatitle {Applying machine learning in science assessment: a systematic review} {Applying machine learning in science assessment: a systematic review}.{\BBCQ}
\newblock
\APACjournalVolNumPages{Studies in Science Education}{56}{1}{111--151,}
\newblock

\newblock

\PrintBackRefs{\CurrentBib}

\end{thebibliography}


\end{document}